\documentclass[12pt, letter]{article}
\usepackage[utf8]{inputenc}
\usepackage[pdftex]{graphicx}
\usepackage[lofdepth,lotdepth,labelformat=empty]{subfig}

\usepackage{rotating}
\usepackage{pslatex}
\usepackage{colortbl}
\usepackage[margin=1in]{geometry}
\usepackage[charter]{mathdesign}
\usepackage{setspace}
\usepackage{lscape}
\usepackage{multirow}
\usepackage{ctable}
\usepackage{amsmath}
\DeclareMathAlphabet{\altmathcal}{OMS}{cmsy}{m}{n}
\usepackage[square,sort&compress,comma,numbers]{natbib}
\usepackage{color}
\usepackage{eurosym}
\usepackage{booktabs}
\usepackage{dcolumn}
\newcolumntype{.}{D{.}{.}{-1}}
\usepackage[toc,page]{appendix}
\usepackage{authblk}
\usepackage{adjustbox}

\usepackage{lineno}

\newcolumntype{C}[1]{>{\centering\let\\\tabularnewline}p{#1}}
\newcolumntype{R}[1]{>{\raggedleft\let\\\tabularnewline}p{#1}}
\newcolumntype{L}[1]{>{\raggedright\let\\\tabularnewline}p{#1}}

\usepackage{helvet}
\usepackage{titlesec}
\titleformat*{\section}{\Large\bfseries\sffamily}
\titleformat*{\subsection}{\large\bfseries\sffamily}

\usepackage{color}
\definecolor{darkred}{rgb}{0.5,0,0}
\definecolor{darkgreen}{rgb}{0,0.5,0}
\definecolor{darkblue}{rgb}{0,0,0.5}
\definecolor{tearose(rose)}{rgb}{0.96, 0.76, 0.76}
\definecolor{gray2}{rgb}{.95,.95,.95}

\usepackage{hyperref}
\hypersetup{colorlinks,
 linkcolor=darkred,
 filecolor=darkred,
 urlcolor=darkred,
 citecolor=darkred,
 pdftitle={multi_comms},
 pdfauthor={author names},
}

\makeatletter
\renewcommand{\maketitle}{\bgroup\setlength{\parindent}{0pt}
\begin{flushleft}
  {\@title}

  \@author
  
\end{flushleft}\egroup
}
\makeatother

\begin{document}
{\sf

\title{\LARGE Multiplex communities and the emergence of international conflict \vspace{.4cm}}

\date{}

\author[1,*]{Caleb Pomeroy}
\author[2]{Niheer Dasandi}
\author[3]{Slava Jankin Mikhaylov}

\affil[1]{Department of Political Science, The Ohio State University, Columbus, Ohio, United States}
\affil[2]{School of Government, University of Birmingham, Birmingham, United Kingdom}
\affil[3]{Data Science Lab, Hertie School, Berlin, Germany}
\affil[*]{pomeroy.38@osu.edu}

\maketitle

\section*{Abstract}

\noindent
Advances in community detection reveal new insights into multiplex and multilayer networks. Less work, however, investigates the relationship between these communities and outcomes in social systems. We leverage these advances to shed light on the relationship between the cooperative mesostructure of the international system and the onset of interstate conflict. We detect communities based upon weaker signals of affinity expressed in United Nations votes and speeches, as well as stronger signals observed across multiple layers of bilateral cooperation. Communities of diplomatic affinity display an expected negative relationship with conflict onset. Ties in communities based upon observed cooperation, however, display no effect under a standard model specification and a positive relationship with conflict under an alternative specification. These results align with some extant hypotheses but also point to a paucity in our understanding of the relationship between community structure and behavioral outcomes in networks. 

}

\begin{spacing}{2}

\section*{Introduction}

Community structure is a fundamental feature of complex networks. The community detection task consists of the identification of subgraphs where vertices exhibit dense within-group ties relative to out-group ties \cite{girvan2002community}. These mesostructural patterns shed light on physical, biological, and social networks, with applications ranging from disease surveillance to paper citations \cite{salathe2010dynamics,calcagno2012flows,menche2015uncovering,lima2015determinants,huttlin2017architecture,strano2018mapping,waniek2018hiding,trujillo2018document}.  Early work on modularity developed a principled assessment of the quality of network divisions \cite{newman2004finding,duch2005community,newman2006modularity}, and the current battery of  detection tools permits investigation of multilayer, multiplex, and time-dependent networks, including algorithms that can accommodate signed edges and heterogeneously structured networks \cite{traag2009community,mucha2010community,benson2016higher,su2017algorithm,wilson2017,zhai2018null}. 

For computational social scientists, this methodological expansion permits investigation of theoretical questions that previously posed modeling challenges at the mesostructural level. The enduring debate on the relationship between interconnectedness and conflict in International Relations (IR) is an exemplary case. On the one hand, Jean-Jacques Rousseau believed that ``...interdependence breeds not accommodation and harmony, but suspicion and incompatibility'' \cite[321]{hoffmann1963rousseau}. More recently, Kenneth Waltz argued that ``the fiercest civil wars and the bloodiest international ones have been fought within arenas populated by highly similar people whose affairs had become quite closely knit together'' \cite[138]{kenneth1979theory}. On the other hand, Immanuel Kant emphasized ``that the growth of interconnectedness demonstrated the existence of the unique human capacity for establishing systems of cooperation...'' \cite{linklater2010global}. More contemporary liberal IR theorists also stress the pacifying effects of interdependence \cite{doyle1986liberalism,oneal1999kantian}.

Empirical investigations of this question typically conceptualize interdependence at the dyad level -- such as trade flow ratios between states $v_i$ and $v_j$ -- and infer relationships to conflict via (generalized) linear models, e.g. \cite{barbieri1996economic,oneal1996liberal}. This conceptualization implies, for example, that if three states $v_i$,  $v_j$, and $v_k$ enjoy a closed triadic cooperation agreement and $v_i$ reneges on the agreement, the exit of state $v_i$ from the commitment to $v_j$ is independent of the exit of state $v_i$ from the commitment to $v_k$. This introduces statistical issues associated with the use of dyads to study $k$-adic phenomena \cite{poast2010mis} and misses the fundamental mechanism of theoretic interest, or as Lupu and Traag \cite[1012]{lupu2013trading} put it: ``...[scholars] have assumed independence in order to study interdependence''. Indeed, it has been suggested that until we ``create and test more complex models, we are not likely to make theoretical progress in sorting out this question'' \cite[56]{mcmillan1997interdependence}.

We draw upon two recent developments relevant to the question of interconnectedness and conflict. First, in international politics, an emerging literature deploys community detection algorithms to examine the role of trade, democracy, and intergovernmental organization dependencies  \cite{lupu2013trading,cranmer2015kantian}, as well as separate attention to alliances \cite{traag2009community} and UN votes \cite{pauls2017affinity}. The common intuition underlying each of these studies is that the community structure of the international system is an underdeveloped predictor of behavioral outcomes. Second, recent findings in the broader network cooperation literature suggest that community structure helps to explain the emergence and maintenance of cooperation on graphs \cite{lozano2008mesoscopic,gianetto2015network} and that multilayer and multiplex structure fosters cooperative stability \cite{gomez2012evolution,wang2012evolution}. These findings are important for network analytic approaches to international politics, because in contrast to laboratory settings with well-mixed populations, states are indeed embedded in multiple layers of potentially interdependent relations. The network cooperation literature, however, has less to say about the relationship between multiplex community structure and other behavioral outcomes, such as conflict. 
  
This paper employs advances in multilayer community detection to locate dense clusters of states and then inferentially models these communities against the emergence of conflict in the international system. Previous work finds pacifying effects of community membership in the traditional Kantian-inspired foci of trade, democracy, and intergovernmental organization networks \cite{lupu2013trading,cranmer2015kantian}. We innovate through attention to data beyond these networks in order to better define the scope of the beneficial effects of community membership on conflict: does the broader cooperative mesostructure of the international system display similar effects, or are previous findings contingent on Kantian-based networks in particular? We consider weaker signals of expressed affinity in the United Nations (UN), as well as stronger signals of observed bilateral cooperation agreements. For the former, we employ layers of UN votes and speeches. For the latter, we search across network layers of science, military, commodity, fishery, and telecommunication cooperation agreements.

The results suggest the following. First, diplomatic cohesion in UN votes and speeches associates negatively with conflict onset. That is, the presence of an affinity community tie in a given dyad correlates with a decrease in conflict likelihood within that dyad. This result provides an extension to a previous finding based upon UN votes alone through the addition of diplomatic speeches in a multilayer setting \cite{pauls2017affinity}. Second, states embedded in cooperation communities appear no more or less likely to engage in conflict under a standard model specification and are more likely to engage in conflict under an alternative specification. This finding contrasts with the often implicit assumption that cooperation community membership reduces the likelihood of conflict amongst members. Furthermore, states who bridge multiple cooperation communities are significantly more likely to experience conflict. These findings lend some support for extant hypotheses but also point to a paucity in current knowledge about the relationship between community structure and behavioral outcomes in social systems.

\section*{Results}
\subsection*{Community detection procedure}

We follow recent work in conceptualizing the international system as a multilayer network \cite{cranmer2015kantian}: a network representation where nodes are connected across layers of different tie sets \cite{kivela2014multilayer,aleta2018multilayer,porter2018multilayer}. Whereas a single mode representation is especially useful for the isolation of specific theoretical mechanisms (e.g. a trade tie's impact on $Y$), we instead aim to capture broader cooperative structure that might exist across layers of the international system. Yet, because innumerable slices of relationships exist in international politics, the resulting communities can quickly become uninterpretable. We therefore focus on two types of multilayer graphs based upon data previously scrutinized by network analysts in IR, namely bilateral cooperation agreements and position affinity expressed in the UN. The former represent stronger signals of observed country-country relations, whereas the latter represent weaker, correlational signals of affinity in expressed preferences.

For strong signal communities, we employ five cooperation topics from the World Treaty Index: science, military, commodities, fisheries, and telecommunications \cite{pearson2001rohn,poast2010electronic}. Previous research finds that network dynamics in part drive bilateral agreement formation and evolution on these topics \cite{kinne2013network}. These topics represent key areas of coordination \cite{krasner1991global,morrow1994modeling} and help to avoid redundancy across layers due to their relative orthogonality. For example, state motivations behind fishery agreement formation differ from motivations behind science agreement formation \cite{haas1980collaborate}. This topical diversity increases confidence that detected communities represent groups of intensive cooperators across issue areas.

For each year, we take the multilayer graph $\altmathcal{G}_t = (\altmathcal{V}, \altmathcal{E}) = \{ G_{t_1}, \dots, G_{t_k} \}$, $i \in \{1, \dots, k \}$  where $G_{t_i} = (V,E)$ is a single elementary network layer that corresponds to one of the five distinct topics. Each layer contains an aligned node set $V = \altmathcal{V}$ with an undirected and unweighted edge $e_{ij} = e_{ji} = (v_i , v_j) \in E$ between nodes $v_i$ and $v_j$ if there exists a bilateral agreement between these two countries in layer $G_{t_i}$. We use a moving window such that an edge is present if a bilateral agreement was initiated within the past ten years, and we assume that the edge dissipates outside of this window. This provides a sequence of yearly multilayer graphs $\altmathcal{S}_{\altmathcal{G}_t} = \{ \altmathcal{G}_1, \dots, \altmathcal{G}_t \}$.

For weak signal communities, we employ UN votes and a recently released dataset of speeches delivered during the annual UN General Debate \cite{baturoetal2016}. UN votes represent a key source of information about the expressed preferences of states \cite{voeten:2000,voeten2004resisting,voeten2013data,bailey2017estimating}. Furthermore, previous network research examines UN voting communities in detail \cite{macon2012community}, including the relationship between community membership and conflict \cite{pauls2017affinity}. In contrast to previous community detection work, however, we employ country ideal points rather than raw UN votes. Noting methodological challenges associated with UN votes, Bailey et al \cite{bailey2017estimating} propose the use of unidimensional ideal points estimated from a dynamic ordinal spatial model. Thus, ideal points derive from a more theoretically-informed model of vote choice given a state's preferences. For each year, we calculate the Euclidean distance between each country pair's ideal points, converting each distance to a similarity score in order to construct a $V \times V$ similarity matrix.

We utilize speeches as the second graph layer in order to align with recent political science research that turns to text data in order to more accurately capture the expressed positions of political actors, e.g \cite{lauderdale2014scaling,kim2018estimating,peterson2018classification}. UN votes often display high cohesion, with states casting votes along regional bloc lines, for ceremonial purposes, or because specific agenda items arise beyond the state's control \cite{voeten2013data,baturoetal2016}. State speeches, on the other hand, provide delegations with greater flexibility to express positions. For example, in 1974 Greece and Turkey voted the most similarly amongst NATO members in the UN General Assembly (with ideal points of 0.68 and 0.42, respectively). Yet, that same year the two country's air forces engaged in a dogfight which led to the death of a Turkish pilot during tensions that arose from Turkey's invasion of Cyprus. In contrast to their votes, their UN General Debate speeches revealed these tensions, with each blaming the other for the crisis. The Supplementary Information (SI) describes this example and others, such as India and Pakistan who engaged in a border conflict in 1999, in greater depth. Thus, the addition of the speech layer helps to capture greater heterogeneity in state positions relative to previous community detection work that focuses on votes alone.

We first embed the speeches into vector space using the Global Vectors for Word Representation (GloVe) algorithm. Word embeddings encode more semantically interesting speech patterns compared to the typical bag-of-words representation of text data \cite{pennington2014glove}. For each year, we utilize the Word Mover's Distance (WMD) in order to locate distances between states' speeches  \cite{kusner2015word}. WMD conceptualizes the state-state speech distance problem as one of minimizing the required effort to move one state's speech embeddings to the vector space location of another state, which we in turn convert to similarity scores \cite{kusner2015word}. This yields a $V \times V$ speech similarity matrix for each year. Because the resultant vote and speech  matrices are densely populated, with each state seemingly connected to every other state, we follow previous work that employs mutual $k$-nearest neighbor graph clustering to yield candidates for multilayer community detection \cite{ozaki2011using,pauls2017affinity}. The notation for the sequence of multilayer weak signal graphs is identical to the bilateral agreements outlined above.

With these strong and weak signal candidate layers in hand, we set about detecting multiplex communities. In international politics, different layers might exhibit heterogenous structure. As mentioned, states might initiate bilateral agreements for topic-dependent reasons, and the vote and speech matrices in Fig 1(A) exhibit heterogenous similarity structures. Most community detection methods, however, posit the same community structure across network layers. Therefore, we employ a newly developed method that can accommodate heterogenous structure, namely the Multilayer Extraction procedure \cite{wilson2017}. The algorithm identifies densely connected vertex-layers in multilayer networks through a significance-based score that quantifies the connectivity of an observed vertex-layer set by comparison with a fixed degree random graph model. The introductory paper provides technical details \cite{wilson2017}.

\vspace{.8cm}

\begin{center}[Fig. 1 about here]\end{center}

\noindent
\textbf{Fig. 1: Multilayer community detection procedure.} In Fig 1(A), mutual 5-nearest neighbor graph clustering on yearly speech (top) and ideal point (bottom) similarity matrices yields candidate adjacency layers for multilayer community detection. Then, we project the edge list recovered from the multilayer extraction algorithm into a single mode network of detected communities. Here, the year 1973 serves as an illustration. The procedure is identical for communities based on cooperation agreements, less the nearest neighbor clustering, since the data are already in adjacency matrix form. Figs 1(B) and 1(C) display the number of detected communities over time for weak and strong signal communities, respectively. Point weights represent the percentage of states that belong to at least one community. Point shading represents the percentage of states that serve as bridges across at least two communities. Note that these results represent the average of the different preprocessing and parameter settings examined. For ease of trend visualization, the plots include a local weighted regression curve.
\vspace{.8cm}

Community detection on yearly instances of strong and weak multilayer networks yields separate sequences of detected community memberships. Single-mode projections of these memberships produce strong and weak multiplex communities for each year, formally $\altmathcal{M}_{strong} = \{ M_{strong_{1970}}, \dots, M_{strong_t} \}$ and $\altmathcal{M}_{weak} = \{ M_{weak_{1970}}, \dots, M_{weak_t} \}$, $t \in \{1970, \dots , 1990 \}$, with ties weighted by the number of common communities between two states. The year 1970 represents the beginning of the sequence, because this is the first available year in the corpus of speeches. The year 1990 serves as the final year in the sequence, because previous international conflict research finds evidence that the structural changes associated with the end of the Cold War led to changes in the causal processes that underlie conflict \cite{jenke2017theme}. Thus, we avoid imposing a model that bridges into the post-Cold War era to avoid the conflation of data generating processes. Furthermore, World Treaty Index data availability declines from the 1990s onwards (see \citep[774]{kinne2013network}).

Fig 1(A) presents the pipeline for the Multilayer Extraction procedure. Figs 1(B) and 1(C) display the number of detected weak and strong signal communities over time, respectively. Point weights indicate the percentage of states that belong to at least one community. Point shading indicates the percentage of nodes that bridge at least two communities. These plots provide a novel glimpse into international polarity with respect to the number of clusters in the system and the ties within and across clusters \cite{de1978systemic}. Larger points and larger numbers of communities suggest a system in which states are more exhaustively divided into groups (i.e. poles). Lighter points indicate a more modular system with fewer bridging ties (i.e. a system that is more polarized given the constellation of poles). The communities detected from cooperation agreements suggest that states are less exhaustively divided into clusters towards the end of the Cold War, evidenced by a decline in the number of communities and a smaller percentage of states assigned to a community. The communities detected from signals of diplomatic affinity suggest a mean increase in the number of communities over time, with a relatively steady and large percentage of states assigned to a community. Further, greater heterogeneity exists in the weak signal graphs, evidenced by a consistently higher number of communities relative to cooperation agreements over time.

\subsection*{The emergence of interstate conflict}

The detected multiplex communities represent the following. The communities based upon stronger signals represent tightly-knit groups of cooperators, taking into account the relational structure at each layer of the multilayer cooperation network. The communities based upon weaker signals represent clusters of states that exhibit similar expressed preferences in the UN, taking into account the similarity structure in the speech and voting layers. Thus, these multiplex communities provide a useful description of the cooperative mesostructure of the international system.

We now investigate the relationship between these communities and the onset of violent conflict in IR. We first consider the effect of community ties at the system level. Then, we restrict the node set to only the most active states in the system to investigate the ways in which different structural roles within these communities correlate with conflict onset. Fig 2 provides a stylized representation of the tie- and node-level effects under consideration.

\vspace{.8cm}

\begin{center}[Fig. 2 about here]\end{center}

\noindent
\textbf{Fig. 2: Conflict effects.} (A) considers the relationship between a community tie and conflict onset at the system level. (B) considers the effect associated with disjoint community membership and conflict onset, i.e. nodes within the same community that lack membership in other communities.  (C) considers the relationship between bridging nodes and conflict onset, i.e. nodes with membership in more than one community.
\vspace{.8cm}

As noted in the Introduction, the networked nature of IR often implies a nonindependence of observations that renders logistic regression unsuitable \cite{cranmer2012complex}. To circumvent these inferential challenges, we employ a temporal extension to the exponential random graph model [(T)ERGM] \cite{robins2001random,hanneke2010discrete}. ERGMs are generative models for network data \cite{wasserman1996logit}, and their results can be interpreted similarly to coefficients from logistic regression: the coefficients provide an estimate for the change in the log-odds likelihood of observing a tie given a one unit change in the independent variable. The outcome network of interest is a yearly snapshot of the conflict onset network. An undirected tie between two states $v_i$ and $v_j$ exists if conflict was initiated in a given year. Model 1 follows a specification by Pauls \& Cranmer \cite{pauls2017affinity} that contains a battery of covariates traditionally associated with conflict onset. This provides a baseline specification and brings our results into proximity with extant findings. The weak and strong multiplex communities then enter the model as an edge-level covariate in Models 2 and 3, respectively. Table \ref{table:tergms1} presents these system-level results.

\begin{table}[!h]
\caption{\textbf{TERGMs: Analysis of international conflict onset, 1970-1990}}
\begin{center}
\begin{adjustbox}{width=.8\textwidth}
\begin{tabular}{l c c c c }
\toprule
 &\bf Model 1 &\bf Model 2 &\bf Model 3 &\bf Model 4 \\
 & Baseline & Weak & Strong & Strong \\
 &&&&(No Contig.)\\
\midrule
Edges                            & $\mathbf{-7.76} $ & $\mathbf{-7.72}$ & $\mathbf{-7.69}$ & $\mathbf{-7.38}$ \\
                                 & $[-8.03;\ -7.49]$    & $[-8.09;\ -7.42]$    & $[-7.98;\ -7.44]$    & $[-7.68;\ -7.14]$    \\
                                 
\hspace{-6mm}\footnotesize{\textit {Multiplex Comms.}} &&&&\\  
Tie Structure			         & \cellcolor{gray2} & \cellcolor{gray2} $\mathbf{-0.60}$ & \cellcolor{gray2} $ \cellcolor{gray2} 0.01$ & \cellcolor{gray2}  $\mathbf{0.41}$  \\
                                 & \cellcolor{gray2} & \cellcolor{gray2} $[-1.36;\ -0.08]$ & \cellcolor{gray2}  $ \cellcolor{gray2} [-0.26;\ 0.26]$ & \cellcolor{gray2} $[0.13;\ 0.73]$      \\
                                 
\hspace{-6mm}\footnotesize{\textit {Network Effects}} &&&&\\                                  
Alternating 2-Stars                         & $\mathbf{1.00}$  & $\mathbf{1.04}$  & $\mathbf{1.02}$  & $\mathbf{0.96}$  \\
                                 & $[0.85;\ 1.13]$      & $[0.83;\ 1.19]$      & $[0.86;\ 1.15]$      & $[0.81;\ 1.09]$      \\
4-Cycles                         & $\mathbf{0.55}$  & $\mathbf{0.56}$  & $\mathbf{0.54}$  & $\mathbf{0.48}$  \\
                                 & $[0.46;\ 0.99]$      & $[0.45;\ 1.13]$      & $[0.44;\ 0.83]$      & $[0.39;\ 0.77]$      \\
GWESP (0)                        & $\mathbf{-0.44}$ & $\mathbf{-0.47}$ & $\mathbf{-0.42}$ & $-0.26$              \\
                                 & $[-1.07;\ -0.18]$    & $[-5.25;\ -0.14]$    & $[-1.05;\ -0.14]$    & $[-0.84;\ 0.05]$     \\

\hspace{-6mm}\footnotesize{\textit {Traditional Covariates}} &&&&\\ 
Joint Democracy                  & $-0.15$              & $-0.16$              & $-0.13$              & $\mathbf{-0.77}$ \\
                                 & $[-0.57;\ 0.24]$     & $[-0.58;\ 0.22]$     & $[-0.56;\ 0.27]$     & $[-1.30;\ -0.32]$    \\
Direct Contiguity				 & $\mathbf{3.78}$  & $\mathbf{3.65}$  & $\mathbf{3.73}$  &                      \\
                                 & $[3.47;\ 4.15]$      & $[3.30;\ 4.07]$      & $[3.43;\ 4.10]$      &                      \\
Capabilities Ratio     		     & $\mathbf{-0.12}$ & $\mathbf{-0.10}$ & $\mathbf{-0.11}$ & $\mathbf{-0.12}$ \\
                                 & $[-0.20;\ -0.07]$    & $[-0.19;\ -0.03]$    & $[-0.19;\ -0.03]$    & $[-0.20;\ -0.05]$    \\
Trade Dependence      			 & $\mathbf{-0.37}$ & $-0.26$              & $\mathbf{-0.39}$ & $0.25$               \\
                                 & $[-1.17;\ -0.06]$    & $[-1.10;\ 0.04]$     & $[-1.25;\ -0.06]$    & $[-0.03;\ 0.40]$     \\
Security IGO Dependence          & $\mathbf{-0.26}$ & $\mathbf{-0.22}$ & $\mathbf{-0.24}$ & $\mathbf{0.18}$  \\
                                 & $[-0.43;\ -0.14]$    & $[-0.38;\ -0.10]$    & $[-0.40;\ -0.12]$    & $[0.07;\ 0.27]$      \\
Economic IGO Dependence	         & $0.00$               & $0.00$               & $-0.01$              & $\mathbf{0.05}$  \\
                                 & $[-0.02;\ 0.02]$     & $[-0.02;\ 0.03]$     & $[-0.03;\ 0.01]$     & $[0.03;\ 0.08]$      \\

\\
Memory (AR, lag = 1)             & $\mathbf{2.97}$  & $\mathbf{2.97}$  & $\mathbf{3.01}$  & $\mathbf{4.39}$  \\
                                 & $[2.62;\ 3.31]$      & $[2.58;\ 3.36]$      & $[2.64;\ 3.36]$      & $[4.16;\ 4.66]$      \\

\bottomrule
\multicolumn{5}{p{.9\linewidth}}{\small{Coefficients in bold are significant at or below the $p$ = 0.05 level. Confidence intervals in brackets are obtained from 2,000 bootstrapped pseudolikelihood replications. Results represent the average of multiple models fitted using a range of robustness checks. All TERGMs run using the {\bf \texttt{\footnotesize btergm}} package \cite{leifeld2015temporal} in the {\bf \texttt{\footnotesize R}} statistical programming environment \cite{R-Core-Team:2017aa}.}}
\end{tabular}
\end{adjustbox}
\label{table:tergms1}
\end{center}
\end{table}

The coefficient sizes and directions are substantively reasonable. The edges term can be interpreted akin to the intercept term in a logit model. For example, the probability of observing conflict within a given dyad is approximately 0.0004 in Model 1. The significance and coefficient direction of the endogenous network statistics of alternating 2-stars, 4-cycles, and geometrically weighted edgewise shared partners (GWESP) indicates that conflict tends to cluster within the network. Further, traditional IR covariates display expected signs and effect sizes. For example, two contiguous states  display a ceteris paribus 3.78 times higher log odds of conflict onset relative to two non-contiguous states, i.e. an odds increase of 43.82.

In Model 2, the coefficient on ties in weak signal communities is significant and negative. This indicates that conflict is less likely between countries that display strong cohesion in their votes and speeches. Specifically, a given dyad's log-odds of experiencing conflict decreases by -0.60 for each additional weak signal community tie within that dyad, all else equal. In Model 3, the coefficient on ties in strong signal communities fails to reach significance. This implies that states with ties in the multiplex cooperation network are no more or less likely to engage in conflict than states without cooperative ties. Model 4 presents a more parsimonious specification (i.e. the omission of direct contiguity) in order to examine the effect of these strong community ties if one were to be observed. The omission of direct contiguity is also intuitive to the extent that cooperation agreements encode regional dynamics (e.g. telecommunication agreements often include neighboring countries), and thus the two variables might compete to explain variance. Under this specification, the cooperation community ties become significant and positive. This finding would indicate that a given dyad experiences an increase in the likelihood of conflict given the presence of a cooperation community tie (or ties) within the dyad. Although the absence of contiguity in this model leads us to caution against over-interpretation of this result, the finding is consistent with the absence of discernible conflict suppression effects given the presence of cooperation agreements.

With these system-level results in hand, we next investigate the different structural roles that members serve in these communities. This provides a more granular understanding  of the mechanisms through which conflict might emerge and diffuse given the structure of the community.  For this analysis, we use the UN as a pivot point and restrict the node set to only those states who voted and delivered a General Assembly speech in a given year. This criteria helps to identify relatively active states in international politics. We note that the results in Table \ref{table:tergms1} are substantively unchanged by this difference in node set.

Two potential mechanisms are of interest. First, the joint community member effect captures states that are in the same community and no other community. For strong signal communities, these states display the highest levels of cooperative dependency, because they lack ties to states in other communities. For weak signal communities, these states display high levels of intragroup diplomatic affinity and lack appreciable connections to other groups in the UN. Second, the community bridge effect captures states who bridge across more than one community. For strong signal communities, these states are less dependent on any single community but are potentially more vulnerable to conflict due to their exposure to multiple communities. For weak signal communities, these states exhibit relatively pragmatic positions that bridge multiple groups in the UN. Table \ref{table:tergms2} presents these results.

\begin{table}[h!]
\caption{\textbf{TERGMs: Analysis of node effects, 1970-1990}}
\begin{center}
\begin{adjustbox}{width=.52\textwidth}
\begin{tabular}{l c c }
\toprule
 &\bf Model 5 &\bf Model 6 \\
 & Weak & Strong \\
\midrule
Edges                           & $\mathbf{-7.71}$ & $\mathbf{-7.59}$ \\
                                & $[-8.08;\ -7.42]$    & $[-8.02;\ -7.23]$    \\

\hspace{-6mm}\footnotesize{\textit {Node Effects}} &&\\
Joint Comm. Member	            & \cellcolor{gray2}$0.23$               & \cellcolor{gray2}$-0.04$              \\
                                & \cellcolor{gray2}$[-0.01;\ 0.46]$     & \cellcolor{gray2}$[-0.26;\ 0.17]$     \\
Comm. Bridge				    & \cellcolor{gray2}$-0.20$              & \cellcolor{gray2}$\mathbf{0.60}$   \\
                                & \cellcolor{gray2}$[-0.70;\ 0.21]$     & \cellcolor{gray2}$[0.26;\ 1.09]$        \\

\hspace{-6mm}\footnotesize{\textit {Network Effects}} &&\\ 
Alternating 2-Stars             & $\mathbf{1.06}$  & $\mathbf{1.08}$  \\
                                & $[0.81;\ 1.23]$      & $[0.87;\ 1.23]$      \\
4-Cycles                        & $\mathbf{0.54}$  & $\mathbf{0.53}$  \\
                                & $[0.44;\ 1.24]$      & $[0.43;\ 1.05]$      \\
GWESP (0)                       & $\mathbf{-0.46}$ & $\mathbf{-0.45}$ \\
                                & $[-5.24;\ -0.12]$    & $[-5.20;\ -0.12]$    \\

\hspace{-6mm}\footnotesize{\textit {Traditional Covariates}} &&\\ 
Joint Democracy                 & $-0.18$              & $-0.24$              \\
                                & $[-0.59;\ 0.17]$     & $[-0.70;\ 0.15]$     \\
Direct Contiguity 			    & $\mathbf{3.67}$  &  $\mathbf{3.72}$                    \\
                                & $[3.33;\ 4.12]$      &  $[3.39;\ 4.14]$                    \\
Capabilities Ratio			    & $\mathbf{-0.11}$ &   $\mathbf{-0.12}$                   \\
                                & $[-0.20;\ -0.03]$    &   $[-0.21;\ -0.05]$                   \\
Trade Dependence			    & $-0.29$              &   $\mathbf{-0.41}$                   \\
                                & $[-1.25;\ 0.03]$     &   $[-1.50;\ -0.02]$                   \\
Security IGO Dependence         & $\mathbf{-0.23}$ &   $\mathbf{-0.15}$                   \\
                                & $[-0.40;\ -0.10]$    &   $[-0.29;\ -0.04]$                   \\
Economic IGO Dependence	        & $-0.00$              &   $-0.02$                   \\
                                & $[-0.03;\ 0.02]$     &   $[-0.04;\ 0.01]$                   \\

\\
Memory (AR, lag=1)              & $\mathbf{2.78}$  & $\mathbf{2.84}$  \\
                                & $[2.43;\ 3.14]$      & $[2.32;\ 3.29]$      \\

\bottomrule
\multicolumn{3}{p{.7\linewidth}}{\scriptsize{Coefficients in bold are significant at or below the $p$ = 0.05 level. Confidence intervals in brackets are obtained from 2,000 bootstrapped pseudolikelihood replications. Results represent the average of multiple models fitted using a range of robustness checks. All TERGMs run using the {\bf \texttt{\footnotesize btergm}} package \cite{leifeld2015temporal} in the {\bf \texttt{\footnotesize R}} statistical programming environment \cite{R-Core-Team:2017aa}.}}
\end{tabular}
\end{adjustbox}
\label{table:tergms2}
\end{center}
\end{table}

For weak signal communities, the results presented in Model 5 indicate a lack of effect for both joint community members and community bridges. This implies that weak community members are no more or less likely to engage in conflict with each other and that bridges are no more or less likely to experience conflict. For strong signal communities, the results of Model 6 suggest a lack of joint community member effect but a significant and positive relationship between conflict and strong community bridges. This implies that states who bridge multiple communities are more likely to experience conflict and perhaps provide a pathway through which conflict might diffuse across communities.

\section*{Discussion}

The above results represent the first evidence on the relationship between multiplex communities and the onset of international conflict beyond previous attention to the Kantian triad (see \cite{cranmer2015kantian}). For communities detected across layers of UN votes and speeches, the results confirm and extend the finding of a previous study based upon voting behavior alone: diplomatic cohesion appears to negatively associate with conflict in the international system \cite{pauls2017affinity}. Although the result is substantively similar, the addition of the speech layer provides useful information on the expressed preferences of states that is otherwise absent in roll call data alone. 

The communities detected across layers of cooperation agreements present a more challenging picture. The most optimistic model specification yields a lack of association between community ties and conflict onset. A more pessimistic specification yields a positive association between cooperative ties and conflict onset. This result is surprising, because cooperation and conflict are often thought to display an inverse relationship, see e.g. \cite{doyle1986liberalism,pinker2012better}. At least two mechanisms might explain this result. First, states at times employ bilateral agreements to manage contentious issues \cite{bueno1981war}. When agreements fail, this tie could provide an indicator for potential conflict onset. Second, those states who interact more often or are most active in agreement formation might face greater opportunities for disputes to arise. Similar arguments have been made in the case of alliance formation and geographically contiguous dyads \cite{gibler1998uncovering,bremer1992dangerous,braithwaite2005location}. For example, Traag and Bruggeman \cite{traag2009community} uncover a similar result in the assessment of their detection algorithm on alliance data, namely that conflict tends to emerge within detected communities. As Waltz \cite[138]{kenneth1979theory} pointed out, ``[i]t is impossible to get a war going unless the potential participants are somehow linked.'' Either way, this finding calls into question the extent to which cooperators enjoy more peaceful outcomes than non-cooperators.  

This communities and conflict puzzle is in part empirically explained by attention to structural roles within communities. Conflict diffusion via network ties is a well-established pattern in IR \cite{zhukov2013choosing}. This study augments this finding: states that bridge cooperative communities are especially conflict prone, and this bridge points to a mechanism through which conflict can diffuse to clusters of states. Those states with exclusive membership in a single community, however, are no more or less likely to engage in conflict with community members. This finding reiterates the open question surrounding interdependence and conflict. Further, this study finds scant evidence that community roles in the UN explain meaningful variance in conflict outcomes: states exclusively aligned with a single bloc and states who pragmatically bridge multiple communities enjoy no detectable change in conflict likelihood. This finding suggests that community membership is more important for conflict outcomes than the specific role that countries serve within communities in the UN. 

Taken together, these results suggest at least two implications. First, for IR cooperation research, increases in tie density do not necessarily lead to decreased levels of conflict. Indeed, previous network science findings indicate that cooperative stability requires enough structure to support cooperation but not so much as to stifle it \cite[2]{gianetto2015network}. Second, for network cooperation research, future work could more rigorously explicate the theoretical mechanisms through which cooperation might suppress conflict. Cooperators tend to cluster on graphs \cite[1561]{nowak2006five}. The above analysis suggests that conflict might diffuse via nodes that bridge these clusters, which could paradoxically increase the likelihood that community members face conflict. Nonetheless, this study's results reiterate the present paucity of observational findings on the relationship between communities and outcomes in social systems. Domain-specific empirical applications will help to narrow the scope of this problem whilst shedding light on the utility of new detection algorithms for questions of computational social science interest.

\section*{Materials and methods}

\subsection*{Data}
As described above, we utilize bilateral cooperation agreements and United Nations (UN) votes and speeches in order to construct the strong and weak signal multilayer graphs, respectively. We obtain the former from the World Treaty Index \cite{pearson2001rohn,poast2010electronic}, which provides the most complete record of bilateral agreements in international relations (IR). These data represent a rich source of information about international cooperation (see e.g. Kinne \cite{kinne2013network}) and have previously been used to operationalize peaceful relations between countries (see e.g. Kasten \cite{kasten2017less}). We specifically include the treaties under the categories of ``Science and Technology'' (7SCIEN), ``Military Procedures'' (9MILIT), ``Raw Materials Trade'' (3COMMO), ``Fisheries'' (8FISH), and ``Telecommunications'' (6TELCO). The dataset contains an edge list of dyads that are party to the treaty, as well as the year that the treaty was signed and a qualitative description of the treaty's purpose. 

For the weak signal data, we employ UN votes and UN General Debate speeches. For roll call data, we utilize yearly country ideal points estimated on a single dimension via a dynamic ordinal spatial model \cite{bailey2017estimating}. This model provides a unidimensional reduction of countries' yea, nay, or abstain decisions on a variety of UN agenda voting items, often interpreted in political science to be a useful indication of a country's expressed preferences or positions with respect to a given topic. The employment of these ideal points helps to avoid the issues posed by the high levels of voting similarity in the UN when attempting to detect communities, as identified in Macon et al \cite{macon2012community}. Furthermore, in contrast to more common unipartite projections of bipartite graphs based on similarity measures (see e.g. Yildim \& Coscia \cite{yildirim2014using}), the ideal points are based on a more explicit theoretical model of vote choice given a state's preferences (see Bailey et al \cite{bailey2017estimating}). These data are available online at Harvard Dataverse: \href{https://hdl.handle.net/1902.1/12379}{hdl:1902.1/12379}. In addition, we utilize the record of annual speeches delivered by country representatives -- predominantly heads of state or government -- during the annual UN General Debate \cite{baturoetal2016}. These speeches are stored as plain text files with associated metadata and are available online at Harvard Dataverse: \href{https://doi.org/10.7910/DVN/0TJX8Y}{doi.org/10.7910/DVN/0TJX8Y}.

The paper's main text describes the vote and speech similarity measures that we employ. In order to move from similarity matrices to candidate adjacency matrix layers for multilayer community detection, we utilize a mutual $k$-nearest neighbor graph approach (see e.g. Ozaki et al \cite{ozaki2011using}). We employ the mutual $k$-nearest neighbor graph approach in order to ensure that our replication procedure follows closely the original clustering procedure of Pauls \& Cranmer \cite{pauls2017affinity}, such that any differences in results can be attributed to the addition of the speeches layer in the multilayer setting. For useful discussions about backboning methods and graph sparsification, see e.g. Serrano et al \cite{serrano2009extracting}, Slater \cite{slater2009two}, and Zhang et al \cite{zhang2014extracting}.

After the performance of community detection on the strong and weak signal graphs, we model the detected communities against the onset of violent conflict in IR. We utilize data from a previous study by Pauls \& Cranmer \cite{pauls2017affinity} that looked at a similar question as the current study, and we thank the authors for sharing these materials. The outcome network of interest is constructed from conflict onset data from the Correlates of War (COW) project's Militarized Interstate Dispute (MID) dataset (v4.1) \cite{palmer2015mid4}. An undirected tie is considered to be present if a MID of level 4 or 5 was initiated between a dyad during the year of interest. These are the two levels of greatest hostility covered in the dataset, with the former corresponding to such actions as occupation of territory or declaration of war, and the latter corresponding to the initiation of war. More details on the conflict data are available online at the Inter-University Consortium for Political and Social Research: \href{https://doi.org/10.3886/ICPSR24386.v1}{doi.org/10.3886/ICPSR24386.v1}.

The inferential model also includes the following covariates. Democracy is a node attribute equal to 1 if the country's Polity IV score is greater than or equal to 7. Direct contiguity enters the model as an indicator variable equal to 1 if two countries share a geographic border or share a sea border within 400 miles of each other. Capabilities ratios capture the ratio of two countries' Composite Index of National Capabilities scores, which utilizes various measures of state capabilities, including population, military expenditures, and iron and steel production. Trade dependence is operationalized as the total yearly trade flow from $v_i$ to $v_j$, divided by the GDP of $v_i$. Finally, security and economic IGO dependence are operationalized as the total number of third-party states to which $v_i$ and $v_j$ are jointly connected through security and economic-oriented intergovernmental organizations, respectively. Pauls and Cranmer \cite{pauls2017affinity} provide more details on these variables.

\subsection*{Models}

To locate vector space representations of the corpus, we utilize the Stanford NLP group's Global Vectors for Word Representation (GloVe) unsupervised learning algorithm \cite{pennington2014glove}. GloVe is a popular log bilinear, weighted least squares model that trains on global word-word co-occurence counts to make efficient use of the corpus statistics. Because it factorizes a word-context co-occurrence matrix, it shares affinities with traditional count methods like latent semantic analysis or principle component analysis. First, the raw texts are stemmed and trimmed of any tokens that appear fewer than 5 times or in fewer than 5\% of speeches across the corpus. This pre-processing was found to improve the quality of the located embeddings. We use a context window of 5 (i.e. 5 words before and 5 words after the target feature). To tune the model's parameters, we fit the model to word vectors of size 50, 100, and 200 with maximum term co-occurrences of 15 and 25 for the weighting function. This yields ``main'' and ``context'' vectors which are subsequently averaged together per the suggestion of the original GloVe paper \cite{pennington2014glove} to locate the final embedding space.

We then calculate the distances between each pair of states in each year using the relaxed variant of the Word Mover's Distance (RWMD) \cite{kusner2015word}. This measure utilizes the embedding space and each country's term-document matrix to measure the cumulative distance required to transform one state's speech point cloud into that of another state. This procedure helps to ensure that distances are not simply a function of the use of different words, but rather differences in the semantic structure of two countries' speeches. The SI presents more details on this procedure. We use the {\bf \texttt{\small quanteda}} package \cite{Benoit2013} for corpus ingestion, and the {\bf \texttt{\small text2vec}} package \cite{text2vec} for fitting the GloVe models and calculating the RWMDs. All analysis is conducted in the {\bf \texttt{\small R}} statistical programming environment \cite{R-Core-Team:2017aa}.

To model the evolution of the conflict onset network, we employ a temporal extension to the exponential random graph model [(T)ERGM] \cite{robins2001random,hanneke2010discrete}. Originally proposed by Wasserman and Pattison \cite{wasserman1996logit} (and also known as $p^{*}$ models), ERGMs are generative models for the performance of inference on network data that have found widespread employment across the network and social sciences \cite{cranmer2010inferential,leifeld2012information,almquist2013dynamic}. The model used here assesses uncertainty using a bootstrap approach proposed by Desmarais and Cranmer \cite{desmarais2010consistent,desmarais2012statistical}, and the models were fitted using the {\bf \texttt{\small btergm}} package \cite{leifeld2015temporal}. Regarding interpretation, our results speak to the likelihood of conflict between two states $v_i$ and $v_j$ given the intensity of cooperation between $v_i$ and $v_j$. We do not extrapolate these results further, such as the likelihood of conflict between $v_i$ and some third party state $v_k$ given the cooperative activity of $v_i$ and $v_j$. At the same time, the results do permit the conclusion that highly active states -- i.e. states with several community ties -- would experience changes in the likelihood of conflict onset commensurate with the number of community ties. See Desmarais \& Cranmer \cite{desmarais2012micro} for more information about the interpretation of ERGMs with respect to various levels of the network.

In addition to the variables outlined above, we specify the following variables in the model. The edges term represents the total number of ties in the graph, akin to the intercept term in regression models. Alternating 2-stars adds alternating sequences of two-paths (i.e. unclosed triangles) to the model, and 4-cycles captures the existence of four nodes connected in a box-like structure, namely $e_{iv} = e_{iu} = e_{jv} = e_{uj} = 1$ \cite{snijders2006new}. Finally, geometrically weighted edgewise shared partners (GWESP) adds a statistic equal to the geometrically down-weighted shared partner distribution, here with a fixed decay parameter of 0. The latter three of these statistics capture potential clustering in the conflict onset network. The community detection results depend on a number of choices surrounding data representation and parameter selection, such as the hyperparameters for the embedding model and the proportion of vertices used to initialize the Multilayer Extraction algorithm. To enhance robustness, we conduct the analysis using the different GloVe hyperparameters described above, as well as vertex initialization proportions of .20, .25, and .30 during the Multilayer Extraction procedure for the strong signal graphs. The results presented in the paper's Emergence of Interstate Conflict section represent the mean results of these analyses.

\subsection*{Data availability}

The datasets and replication code required to produce the results from the analysis are available on Harvard Dataverse at: [link here]

\end{spacing}


\section*{Supporting information}

\doublespacing 

\noindent
{\bf Supplementary text}

\noindent
{\bf Table S1. Nearest features based on cosine similarity.}

\noindent
{\bf Fig S1. t-SNE projection.}

\noindent
{\bf Fig S2. WMD abstract example.}

\noindent
{\bf Figs S3-S5. In-sample goodness-of-fit.}

\noindent
{\bf Fig S6. Test set predictive accuracy.}

\section*{Figures}

\begin{figure}[h!]
\centering
\includegraphics[width=.9\textwidth]{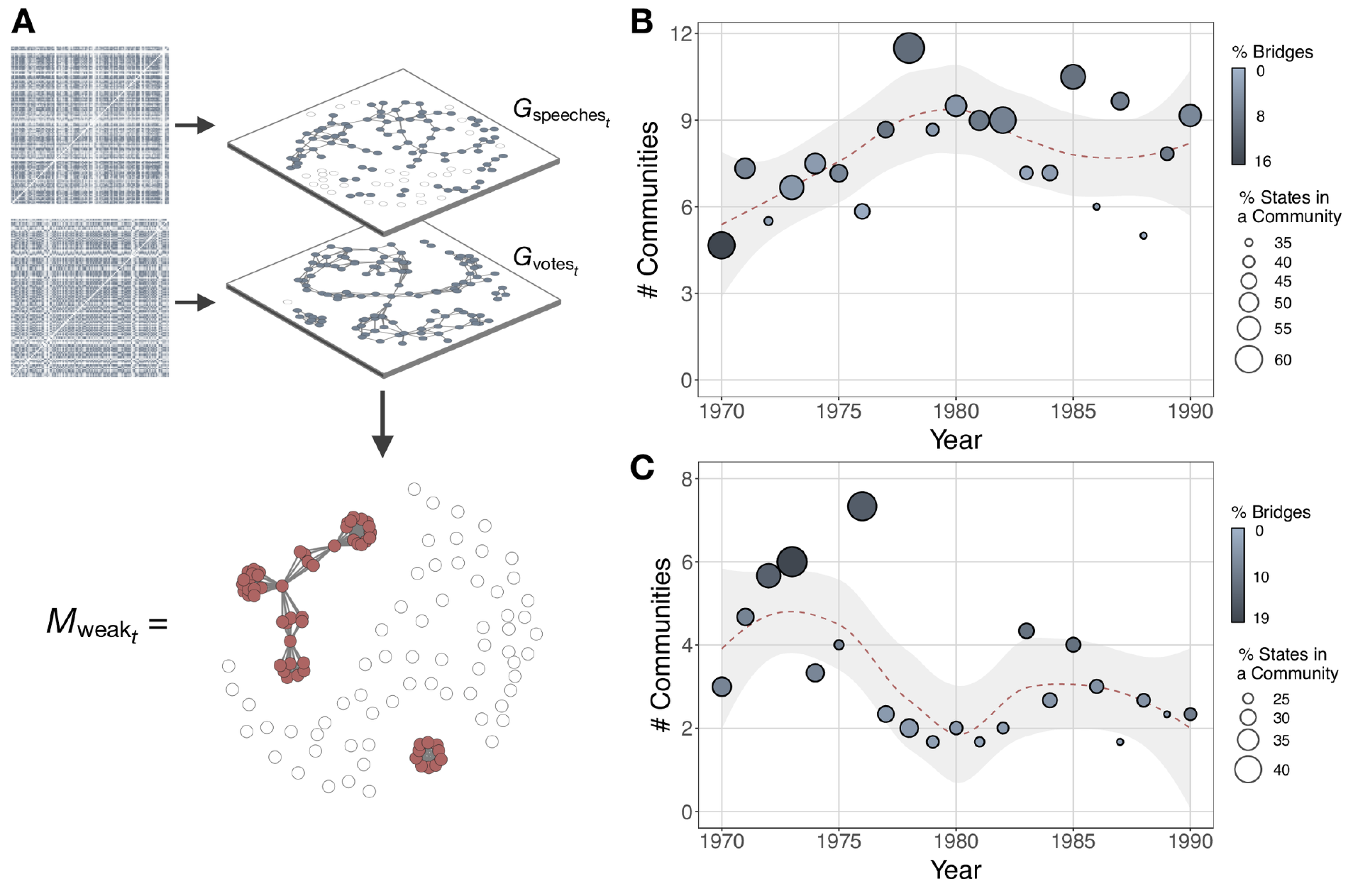} 
\label{fig:fig1}
\caption{\emph{Multilayer community detection procedure.} }
\end{figure}

\begin{figure}[h!]
\centering
\includegraphics[width=.9\textwidth]{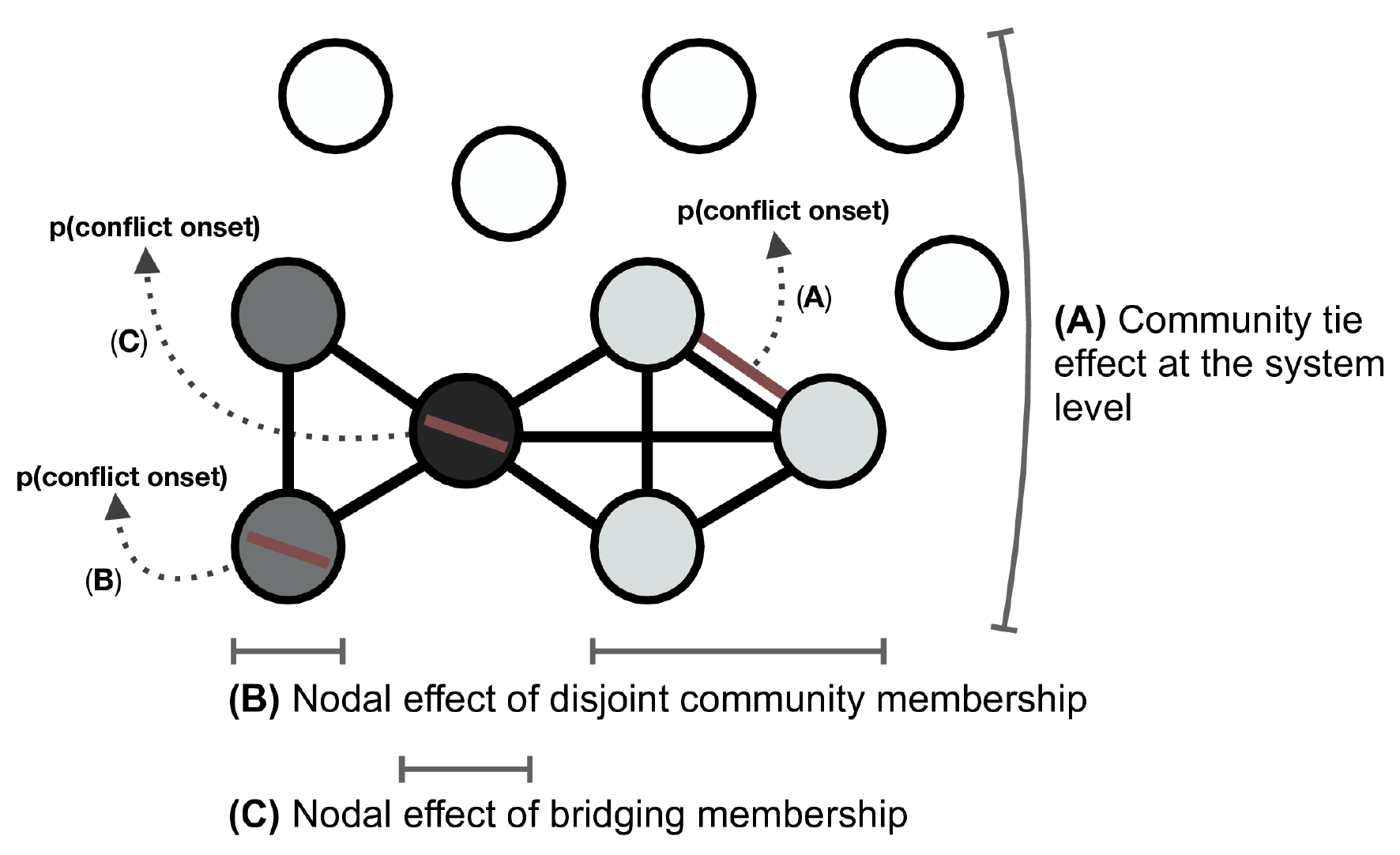} 
\label{fig:fig2}
\caption{\emph{Conflict effects.}}
\end{figure}

\clearpage

\section*{Supporting Information}

\section*{The UN General Debate Corpus}

We draw on the newly released \emph{UN General Debate Corpus} \cite{baturoetal2016} which contains every country statement in the UN General Debate between 1970 and 2017.  The General Debate (GD) takes place every September at the start of each new session of the UN General Assembly (UNGA). It provides all member states with the opportunity to address the UNGA and to present their perspective on key issues in world politics. Governments use their GD statements to put on the record their position on events that have occurred during the past year and on longer-term underlying issues in world politics related to issues such as conflict, terrorism, development, human rights, and climate change. 

A principal difference between GD statements and UNGA voting is that the GD statements are not institutionally connected to decision-making in the UN. As a result, governments are free to discuss the issues they consider to be of greatest importance in world politics, regardless of whether an issue is on the formal agenda of the UNGA. Therefore, as Smith \cite{smith2006} notes, the General Debate acts ``as a barometer of international opinion on important issues, even those not on the agenda for that particular session.'' In providing information about states' preferences on world politics, the GD provides a valuable data source for measuring polarization in International Relations. In addition to being the one major forum where states present their views on international politics free from external constraints, the fact that it takes place annually and includes all UN member states enables comparison over time and across countries. Readers are encouraged to consult Baturo et al \cite{baturoetal2016} for a comprehensive introduction to the corpus. 

As stated in the main text, we discuss an example where disagreement is obvious in states' GD speeches but less obvious in their voting behavior. Consider the following brief excerpts from the GD speeches of Greece and Turkey in 1974.

\begin{quote}
{\bf \small{Greece:}} \textsf{On 15 July a coup, condemned by all of us, was staged to overthrow Archbishop Makarios, the legitimate, elected President of the Republic. This coup was not directed against the Turkish Cypriot community of the island... During the fighting while the coup was in progress, not a single Turkish Cypriot was killed or injured. Yet five days later, large Turkish invasion forces were landing in Cyprus and the Turkish Air Force was launching indiscriminate attacks against unarmed civilians, under the flimsy pretext of protecting the Turkish Cypriot minority on the island, which, I repeat, had not been harmed in any way... Two hours later, the Turkish troops were on the move again, sowing death and destruction, killing United Nations troops, bombing hospitals and schools. Repeated cease-fire calls by the Security Council went unheeded. Turkey even ignored the ceasefire proclaimed by its own Prime Minister on 16 August 1974.} \\

{\bf \small{Turkey:}} \textsf{Turkey has constantly had to face faits accomplis of increasingly serious scope, particularly since 1963. The most recent and the most serious of these faits accomplis was, as we all know, that of 15 July last: a foreign Power undertook a coup d'etat which it had long been fomenting and the purpose of which was to annex the island... The coup d'etat of 15 July was directed precisely against the Turkish community and was directly aimed at the annexation of the island to Greece... I have not, however, finished correcting all the false allegations and baseless charges made by my colleague. I reserve the right to do so when we speak on this matter before the General Assembly. My Greek colleague's speech, unfortunately, shows the nature of the atmosphere in which the debate will take place on the future of the two communities, Turkish and Greek, in the island.} \\
\end{quote}

The two representatives are outlining their positions on the controversy related to the Turkish invasion of Cyprus. Expressed disagreement on this topic is clearly present in the speeches, but as mentioned in the main paper, the two countries' voting ideal points for that year are the most similar amongst all NATO members. A further example is illustrated in the speeches and voting habits of India and Pakistan in 1999, the year the two countries went to war (the Kargil War). Consider the following excerpts from their General Debate statements that year:

\begin{quote}
{\bf \small{Pakistan:}} \textsf{The Kargil crisis was a manifestation of the deeper malaise spawned by the unresolved Kashmir problem and India's escalating repression of the Kashmiri people. India launched a massive military operation in Kargil and
threatened a wider conflict by mobilising its armed forces all along the Pakistan-India international border. Pakistan
acted with restraint... India's repression in Jammu and Kashmir has killed thousands of Kashmiris, forced hundreds of thousands into exile, led to three wars between Pakistan and India and consigned the two countries to a relationship of endemic conflict and mistrust.} \\

{\bf \small{India:}} \textsf{Premeditated aggression by regular forces was committed against India. Not simply was the Lahore Declaration violated, but so was the Simla Agreement, which had prevented conflict for more than a quarter of a century. In self-defence, yet with the utmost restraint, India took all necessary and appropriate steps to evict the aggressor forces from its
territory.... We have been greatly disappointed by this compulsive hostility of Pakistan, because it is an aberration in our region today, where all the other South Asian Association for Regional Cooperation (SAARC) countries are at peace with each other, and are trying, bilaterally and through the SAARC mechanisms, to tackle together the great challenge of development.} \\
\end{quote}

Tensions are clearly present in the textual data of the respective countries. That same year, however, India and Pakistan casted very similar votes in the UN, with ideal points of -0.797 and -0.739, respectively. Therefore, both sources of data appear to provide useful signals of different aspects of underlying state preferences.

\section*{Word embeddings}

In order to use texts together with votes to estimate position affinity, we first consider how to better exploit the information contained in textual data, namely unsupervised learned word embeddings. In the broader natural language processing (NLP) literature, there has been a surge of research devoted to the development of distributional representations of speech which retain syntactical language qualities in ways that the bag-of-words (BOW) approach typically used in political text analysis research is not equipped to retain. The hypothesis claims that words that occur in similar contexts tend to have similar meanings \cite{turney2010frequency}. When operationalized, the unique intuition is that similar words and phrases, such as \emph{ ``atomic,  weapons''} and \emph{``nuclear, warheads''} are found in relatively proximate vector space locations. Although the BOW performs surprisingly well, this example has no features in common, and a BOW representation would assign low similarity scores or high distances. Word embeddings help to ensure that communities are detected amongst states that are actually expressing different positions, as opposed to simply using different language to express the same sentiment. 

When results are projected onto a two dimensional surface, language relationships surface, such as the clustering of synonyms, antonyms, scales (e.g. \emph{democracy} to \emph{authoritarianism}), hyponym-hypernyms (e.g. \emph{democracy} is a type of \emph{regime}), co-hyponyms (e.g. \emph{atomic bombs} and \emph{ballistic missiles} are types of \emph{weapons}), and groups of words which tend to appear in similar contexts like \emph{diplomat}, \emph{envoy}, and \emph{embassy}. Mikolov and collaborators introduce an evaluation scheme based on word analogies that examines dimensions of difference in vector space  \cite{mikolov2013distributed,mikolov2013efficient}. They originally reached the surprising conclusion that simple vector addition and subtraction  uncovers interesting linear substructures of human language, famously that $\mathsf{king} - \mathsf{man} + \mathsf{woman} = \mathsf{queen}$. 

To locate vector space representations of our corpus, we utilize the Stanford NLP group's Global Vectors for Word Representation (GloVe) unsupervised learning algorithm \cite{pennington2014glove}. GloVe is a popular log bilinear, weighted least squares model that trains on global word-word co-occurence counts to make efficient use of the corpus statistics. Because it factorizes a word-context co-occurrence matrix, it is closer to traditional count methods like latent semantic analysis or principle component analysis.\footnote{For recent reviews of the distributional semantics literature, see Turney \& Pantel \cite{turney2010frequency} and Lenci \cite{lenci2017distributional}.} Readers are encouraged to consult the GloVe paper for full technical details, but we describe our approach and resultant vector space here. The model is expressed as:

\begin{equation}\tag{S1}
\mathsf{J(\theta) = \frac{1}{2}\sum_{i,j=1}^{W}}f\mathsf{(P_{ij})(u_i^T v_j - \log P_{ij})^{2}}
\end{equation}

\noindent
where $\theta$ represents parameters, $\mathsf{W}$ is the vocabulary size, $\mathsf{u \in \!R^d}$ and $\mathsf{v \in \!R^d}$ are column and row word vectors, $\mathsf{P_{ij}}$ is the co-occurrence matrix of all pairs of words that ever co-occur, and $f(\cdot)$ is a weighting function which assigns lower weights to words that frequently co-occur. This lattermost term serves as a cap on very frequent words, for example articles like ``the" which provide little predictive information. The algorithm seeks to minimize the distance between the inner product of the word vectors and the log count of the co-occurrence of the two words. Compared to skip-gram approaches which update at each context window, it is clear from the utilization of $\mathsf{P_{ij}}$ that the model trains relatively quickly since it uses the known corpus statistic of word co-occurrences for the entire corpus at once. To our knowledge, this is one of the first IR applications to use word embeddings (see also \cite{lauretig2019identification}).

The Models section of the main paper outlines the parameters we chose.\footnote{As with similar machine learning tasks, we note that our results are sensitive to choices of hyperparameter settings and random seeds (see e.g. Henderson et al \cite{henderson2018deep}).} We follow the computer science literature suggestion of tuning these parameters until reasonable and reliable linear combinations of language are located. Future work should explore in greater detail how systematic tuning decisions for social science applications can be made. Here, we present various qualitative checks on the located embeddings. First, we consider the analogical performance of various features:

\begin{equation}\tag{S2}
\begin{split}
\vec{v}\mathsf{(``peac\textnormal{-}")} \ - \ \vec{v}\mathsf{(``agreement")} \ + \ \vec{v}\mathsf{(``weapon")} \ = \ \vec{v}(\underset{.64}{\operatorname{\mathsf{``nuclear"}}}) \ , \ \vec{v}(\underset{.62}{\operatorname{\mathsf{``destruct"}}}) \\
\vspace{.5cm}
\vec{v}\mathsf{(``west")} \ - \ \vec{v}\mathsf{(``nato")} \ + \ \vec{v}\mathsf{(``russia")} \ = \ \vec{v}(\underset{.57}{\operatorname{\mathsf{``east"}}}) \ , \ \vec{v}(\underset{.50}{\operatorname{\mathsf{``pakistan"}}}) \\
\vspace{.5cm}
\vec{v}\mathsf{(``terrorist")} \ + \ \vec{v}\mathsf{(``bomb")} \  = \ \vec{v}(\underset{.82}{\operatorname{\mathsf{``attack"}}}) \ , \ \vec{v}(\underset{.63}{\operatorname{\mathsf{``barbar\textnormal{-}"}}}) \\
\vspace{.5cm}
\vec{v}\mathsf{(``environment")} \ + \ \vec{v}\mathsf{(``pollut\textnormal{-}")} \  = \ \vec{v}(\underset{.74}{\operatorname{\mathsf{``degrad\textnormal{-}"}}}) \ , \ \vec{v}(\underset{.65}{\operatorname{\mathsf{``ecolog\textnormal{-}"}}}) \\
\end{split}
\end{equation}

\noindent
where each $\vec{v}$ describes a vector space location of the given feature, and the cosine similarity between each vector space location is added or subtracted to find the closest vector offsets (with cosine similarity printed underneath). These analogies are interpreted, for example, as $\mathsf{``agreement"}$ is to $\mathsf{``peace"}$ as $\mathsf{``weapon"}$ is to $\mathsf{``destruct"}$. These examples appear to encode relations of cause-effect and geographic alliance patterns, respectively. The latter two are not analogies, but rather the resultant vector space location when the first two vectors are added together. Although these linear combinations look quite reasonable, it is worth checking the nearest neighbors (as measured by cosine similarity) of various features in the embedding space. These are listed in Table \ref{table:nearest}.

As with the analogical examples, it appears in Table \ref{table:nearest} that reasonable groupings of words are detected. Finally, to get a sense of the semantic structure of the embedding space, we plot the 200 nearest words to the vector space of ``weapon'' and project this onto two dimensions using the common t-SNE algorithm.

\begin{figure}[h!]
\centering
\includegraphics[width=.9\textwidth]{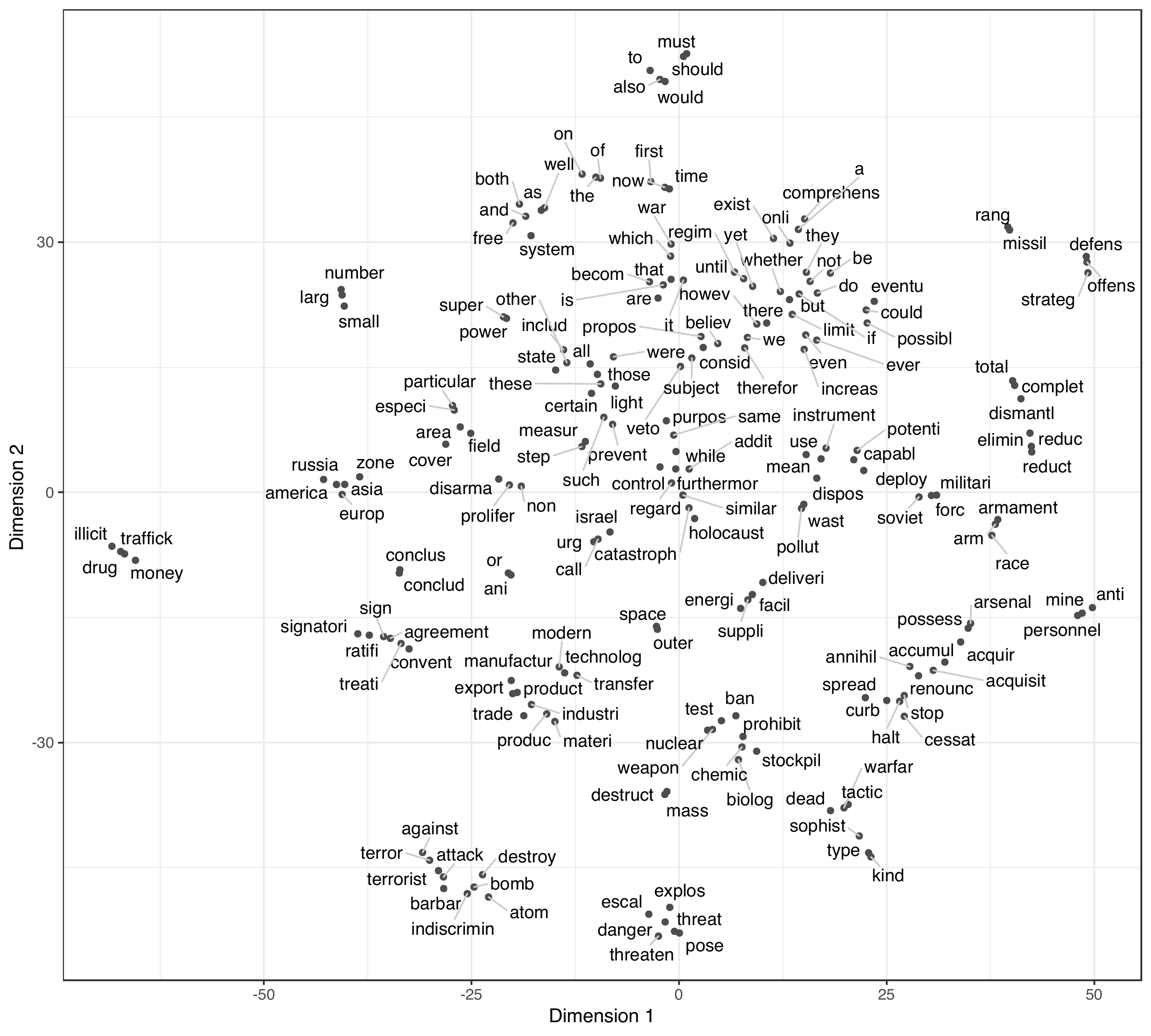} 
\caption{\emph{t-SNE Projection.} 200 nearest words to the vector space of ``weapon'', projected onto two dimensions using the t-SNE algorithm.}
\label{fig:space}
\end{figure}

Fig \ref{fig:space} shows quite intuitive clustering of features. For example, in the third quadrant exists a cluster of features that generally relate to the industrial aspects of weaponry, evidenced by terms like ``manufactur-,'' ``produc-,'' and ``export.'' In quadrant two we see a geographic clustering with terms like ``asia,'' ``europ-,'' and ``america.'' Finally, in the bottom right of quadrant four exists a clustering of terms commonly associated with the international regulation and governance of weaponry, with feature like ``prohibit,'' ``chemic-,'' and ``biolog-.'' These qualitative checks increase our confidence that the located embedding space should contain useful information about the speeches delivered by states. As found in the wider NLP literature, the implication is that these vector space models are surprisingly effective at capturing different lexical relations, despite the lack of supervision. 

\begin{figure}[h!]
\centering
\includegraphics[width=.9\textwidth]{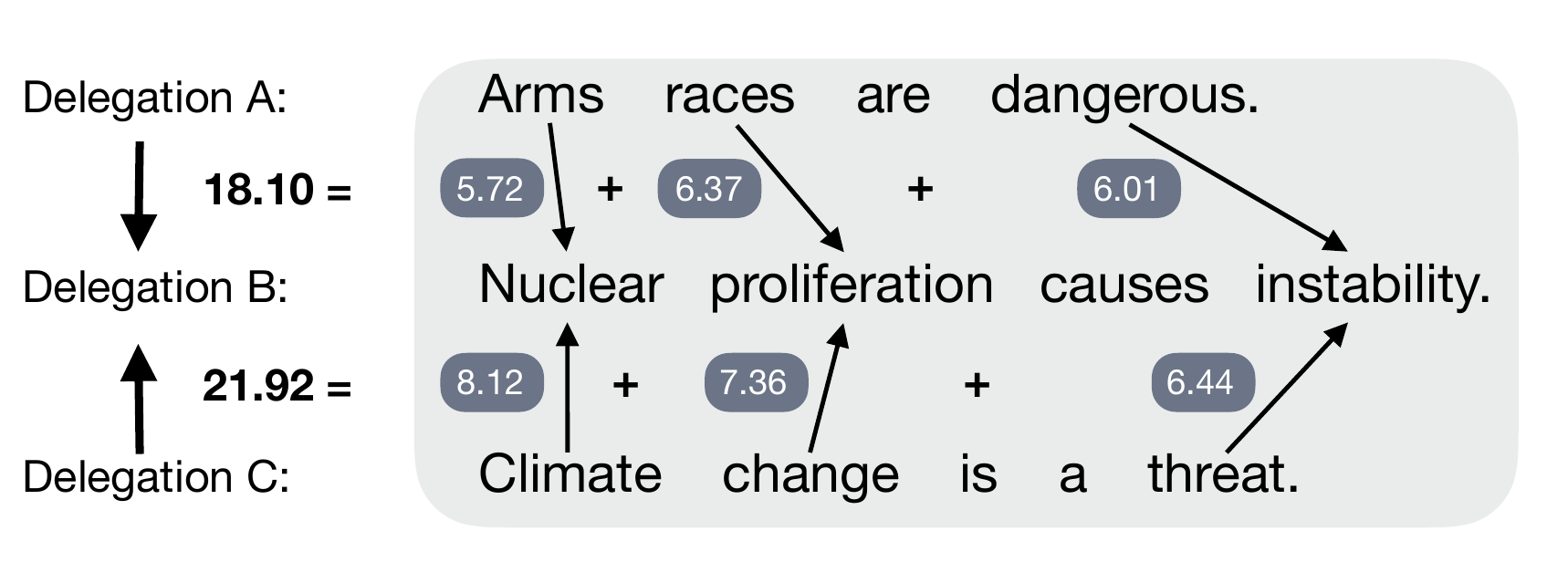} 
\caption{\emph{WMD Abstract Example.}  Three strings of text serve as an example of the motivation behind the employment of WMD. Delegations A and B both discuss issues surrounding international armament but utilize different language. This leads to very low similarity scores under a bag-of-words framework. WMD innovates by capturing the distance required to move one document to the vector space location of another document. In this example, although all three delegations discuss issues relevant to security, the speeches of delegations A and B are nearer in vector space than the speeches of delegations B and C. Actual Euclidean distances from our estimated embeddings are used for illustration. Adapted from Fig 2 in Kusner et al \cite{kusner2015word}.}
\label{fig:wmd_abstract}
\end{figure}

To measure expressed (dis)agreement in these speeches, it is necessary to derive a document-level representation of the learned embeddings.  Although well-established measurements based on cosine similarity, Euclidean distance, or Pearson correlations could be applied to the word embeddings, we utilized the relaxed variant of a relatively newly introduced document distance measure that exploits information contained in both the word embeddings and term-document matrices: the (relaxed) Word Mover's Distance [(R)WMD] \cite{kusner2015word}. WMD measures the cumulative distance required to transform one state's speech point cloud into that of another state, ensuring that differences do not simply reflect the use of different words. States employ varied language and lexical patterns to describe similar topics. For example, if state A says ``nuclear weapons are bad," and state B says ``atom bombs are terrible," the only feature in common is the term ``are," which leads to near-orthogonality in their BOW vectors and low similarity scores. If a third state C says ``atom bombs are good," then B and C would exhibit the highest cosine similarity of the three, despite having the opposite expressed policy positions. Fig \ref{fig:wmd_abstract} shows an illustration of the motivation behind this distance metric, which has been shown to yield state-of-the-art classification accuracy \cite{huang2016supervised}.

Although WMD is relatively fast to compute, we use the relaxed variant (RWMD), which results in tighter bounds and is shown to yield lower test error rates. In short, this relaxes the optimisation problem through the removal of one of the two constraints. If we let $\mathbf{d}$ and $\mathbf{d^{\prime}}$ be the BOW representations of two documents in the $n-1$ dimensional simplex of word distributions which we obtained above, we can express RWMD as:

\begin{equation}\tag{S3}
\mathsf{\min_{\mathbf{T} \geq 0} \quad \sum_{i,j=1}^{n}\mathbf{T}_{ij}c(i,j)} \ \ \ \ \ \ \ \   \mathsf{s.t.} \quad \sum_{j=1}^{n} \mathbf{T}_{ij} = d_i \quad \forall i\in \{1,...,n\}.
\end{equation}

\noindent
where $\mathbf{T} \in \!R^{n \times n}$ is a sparse flow matrix where $\mathbf{T} \geq 0$ denotes how much of word $i$ in $\mathbf{d}$ travels to word $j$ in $\mathbf{d^{\prime}}$ and $\mathsf{\sum_{i,j=1}^{n}\mathbf{T}_{ij}c(i,j)}$ represents the distance between the two documents, i.e. the cost of moving all words from $\mathbf{d}$ to $\mathbf{d^{\prime}}$. Then, the optimal solution is found when each word in $\mathbf{d}$ moves all of its probability mass to the most similar word in $\mathbf{d^{\prime}}$. This optimal matrix $\mathbf{T}_{ij}^{\ast}$ is decided by:

\begin{equation}\tag{S4}
 \mathbf{T}_{ij}^{\ast} =
\begin{cases}
    d_i    & \quad \mathsf{if} \ j = \mathsf{argmin_{j}}  c(i,j) \\
    0    & \quad \mathsf{otherwise.}\\
\end{cases} 
\end{equation}

\noindent
where $d_i$ is the distance of interest which we normalize and convert to a similarity score. The result is a list of $V \times V$ matrices $\mathbf{A}$ with one matrix $A_t$ for each year in the corpus and where $A_{ij}$ is the speech similarity score between states $i$ and $j$ with the diagonals of the matrices set to 0. The original paper provides further details \cite{kusner2015word}.

\section*{Model evaluations}

\begin{landscape}
\begin{figure}[h!]
\centering
\includegraphics[width=.46\textwidth]{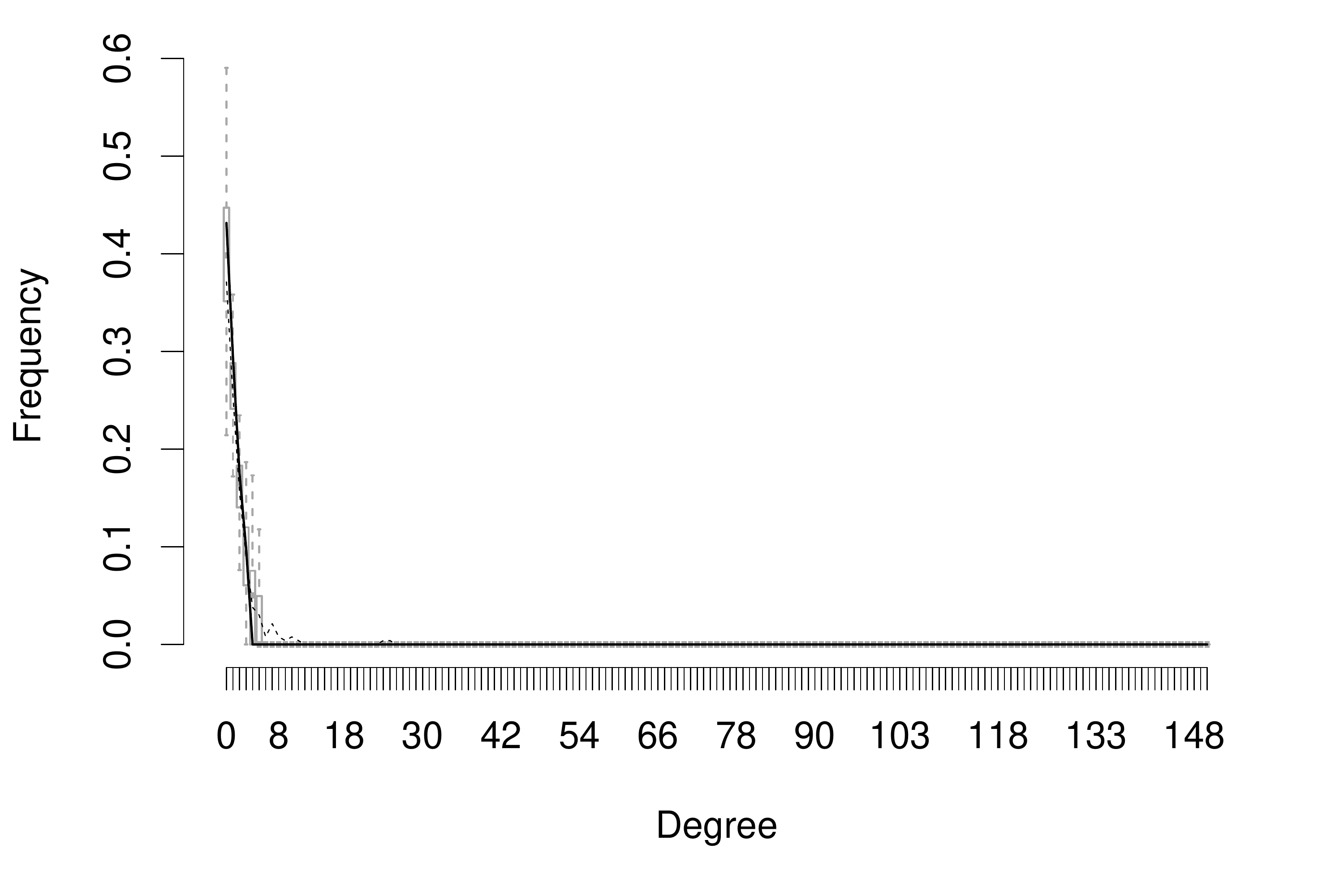} 
\includegraphics[width=.45\textwidth]{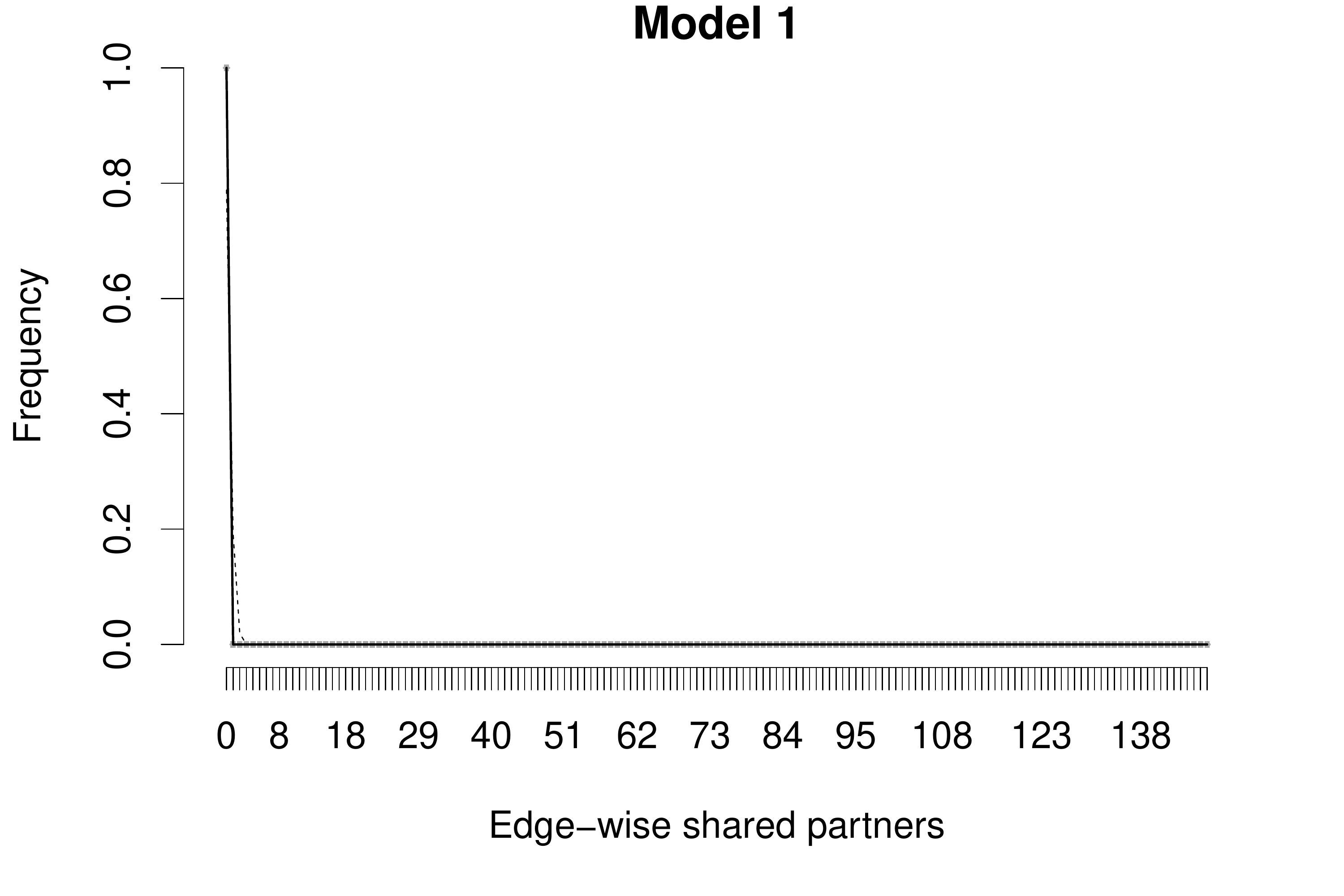} 
\includegraphics[width=.46\textwidth]{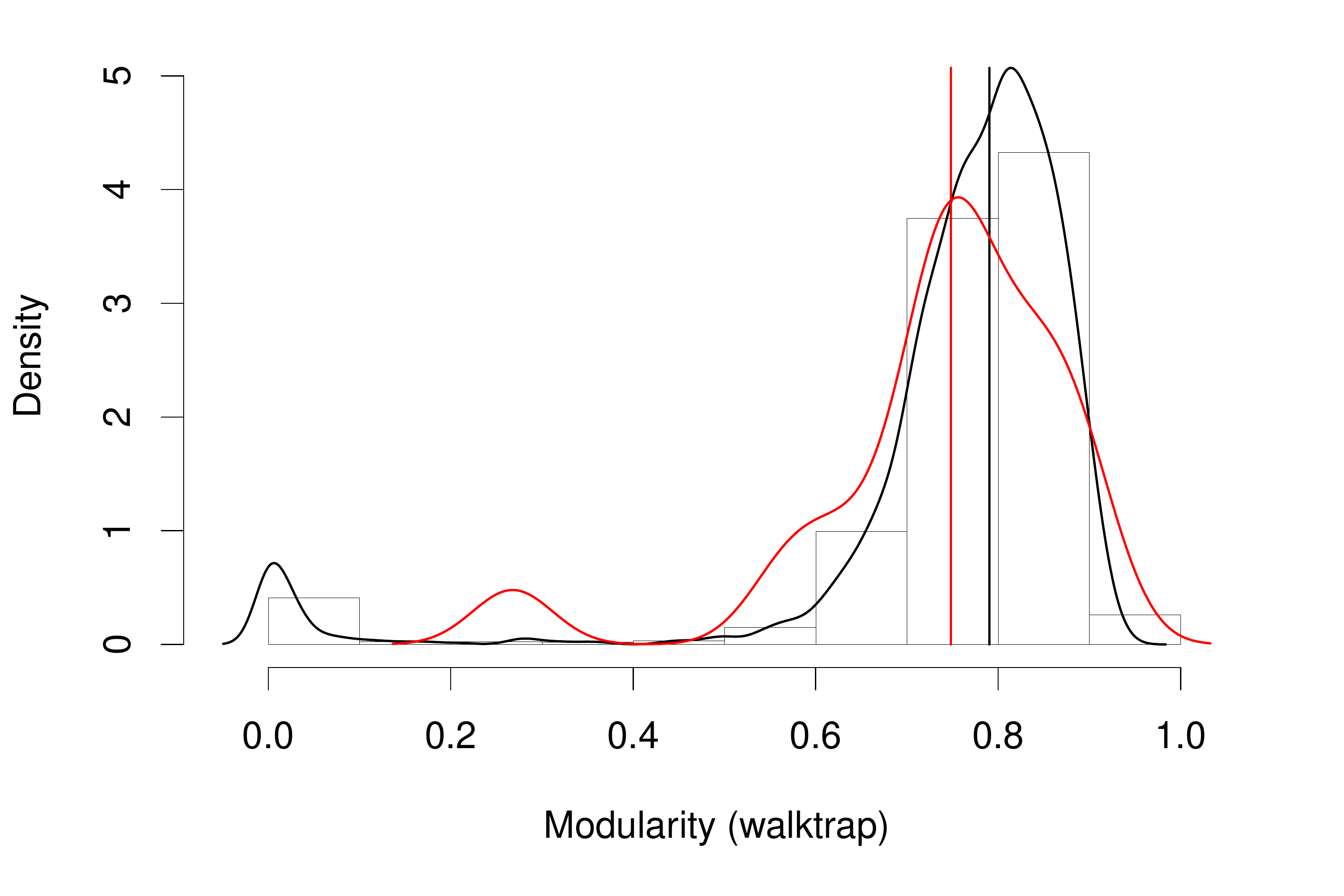} 
\includegraphics[width=.46\textwidth]{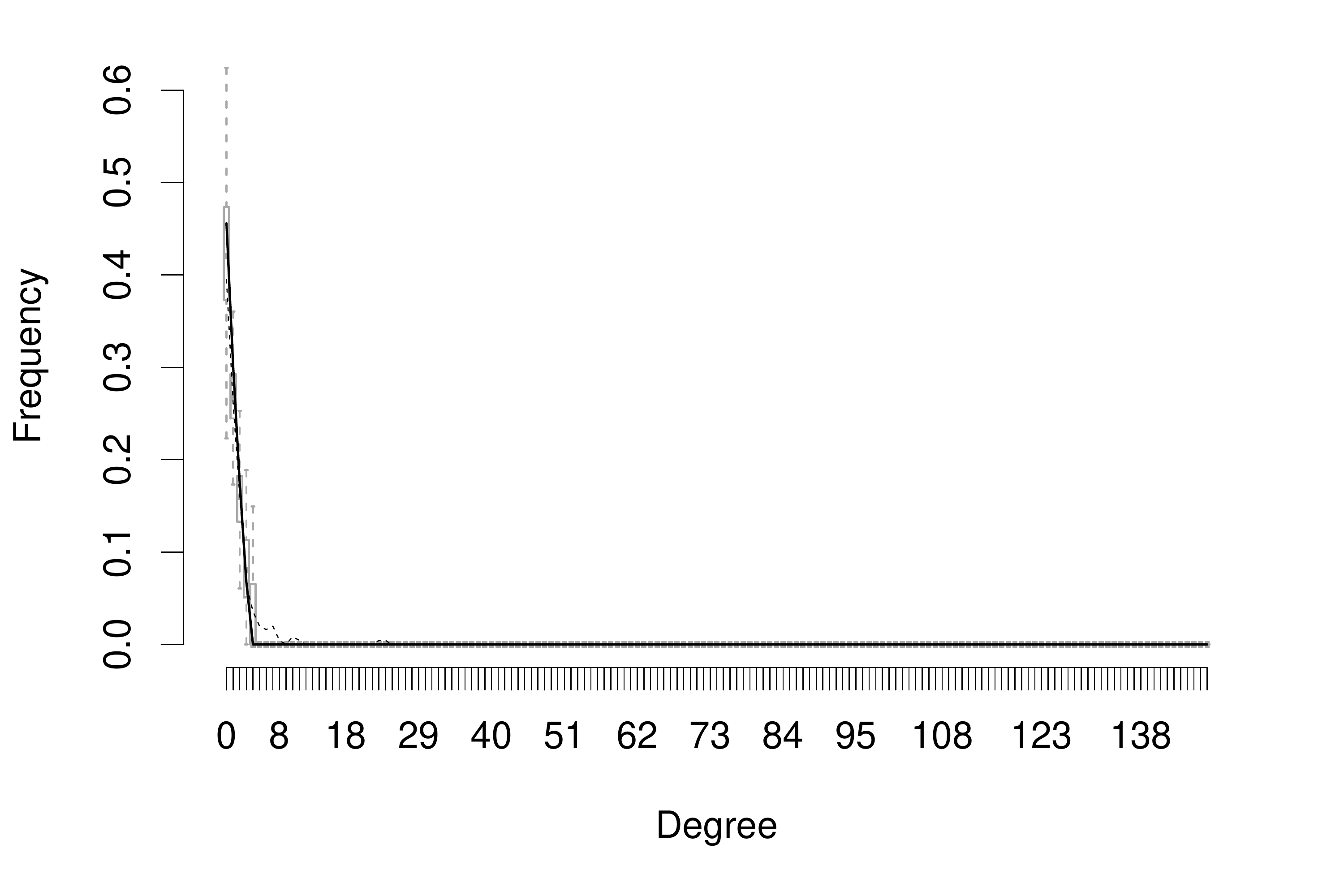} 
\includegraphics[width=.45\textwidth]{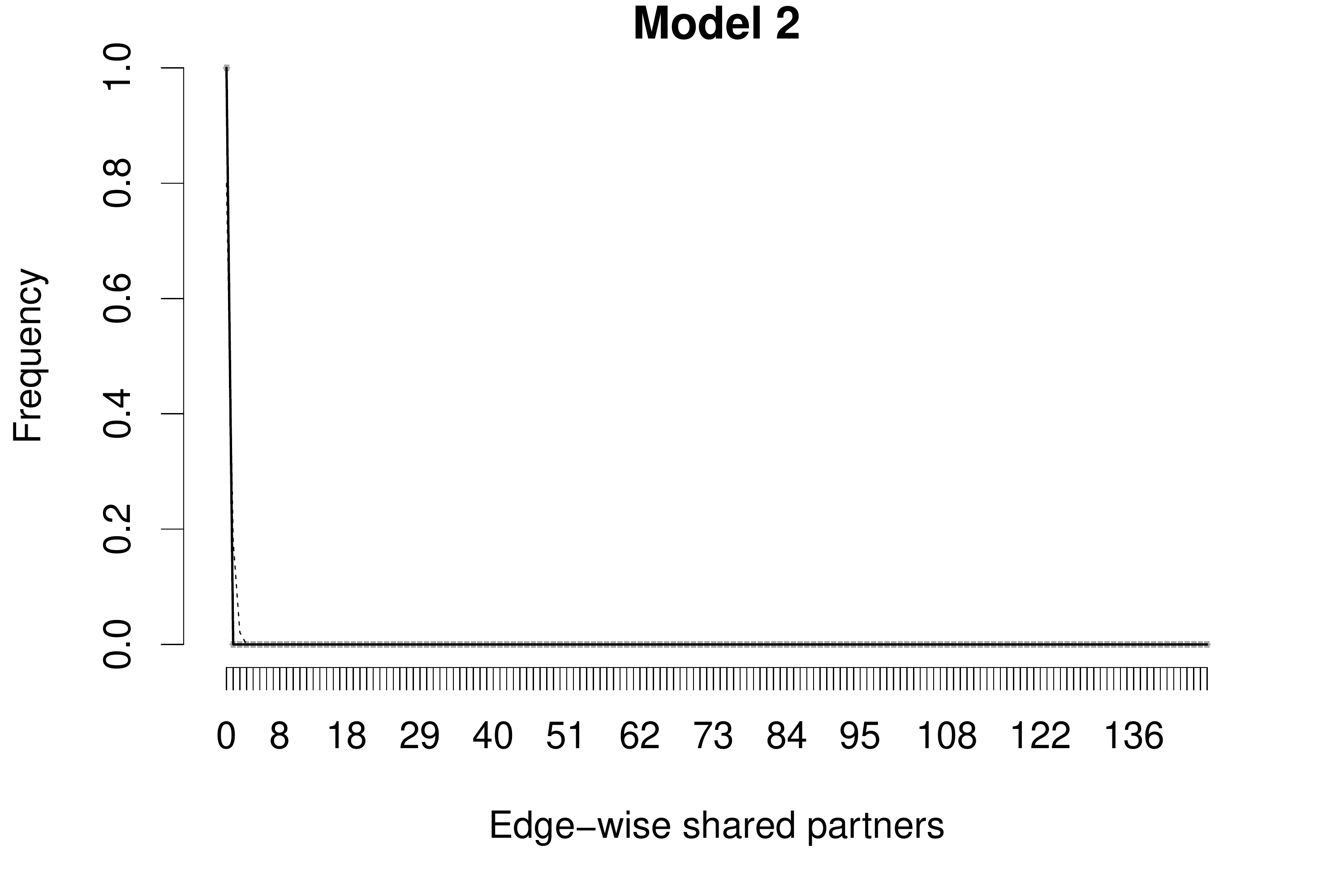} 
\includegraphics[width=.46\textwidth]{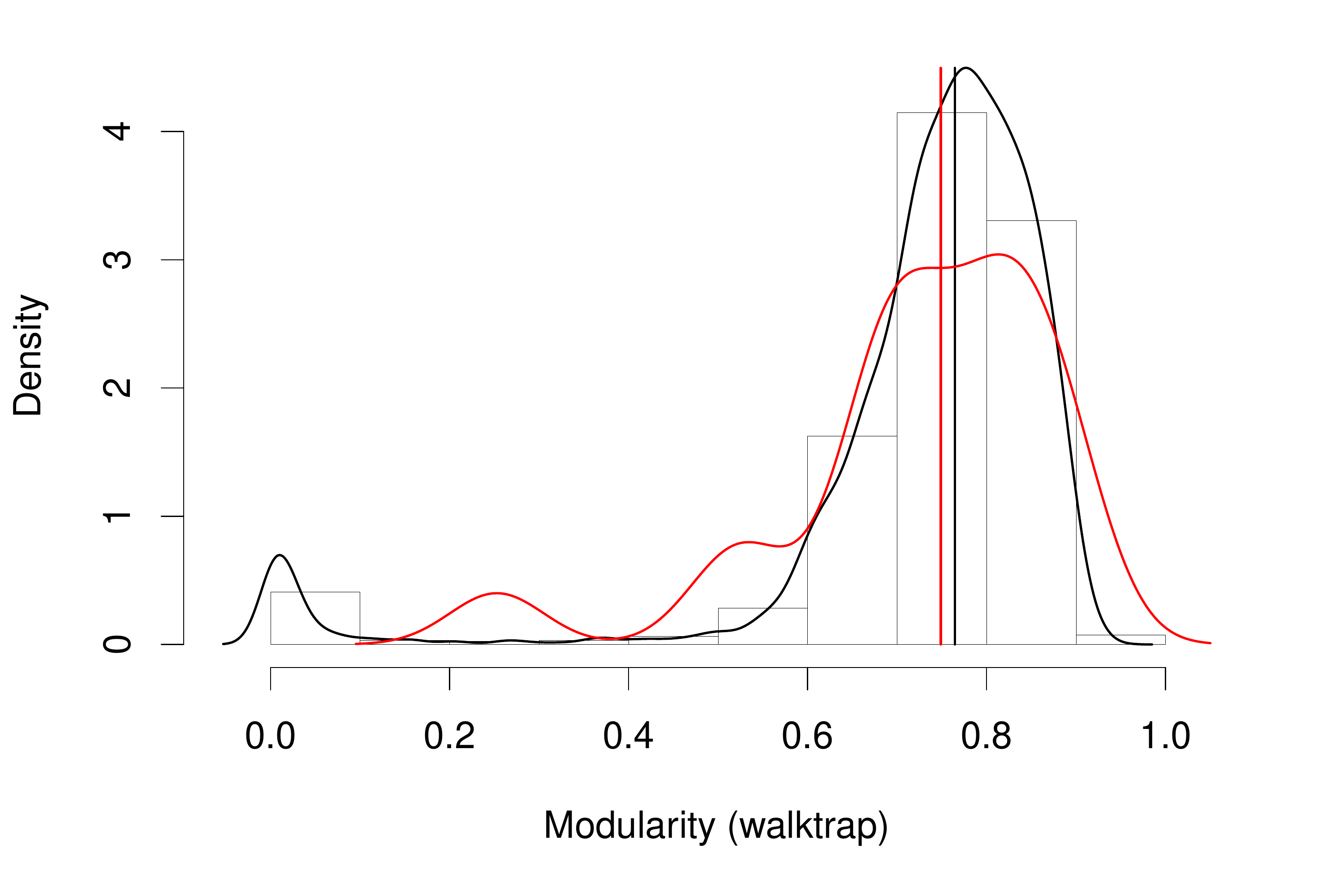} 
\caption{\emph{In-Sample Goodness-of-Fit.} In-sample goodness-of-fit diagnostics as measured by degree, edge-wise shared partners, and modularity, for Model 1 (top row) and Model 2 (bottom row) from the main paper.}
\label{fig:gof1}
\end{figure}
\end{landscape}

\begin{landscape}
\begin{figure}[h!]
\centering
\includegraphics[width=.46\textwidth]{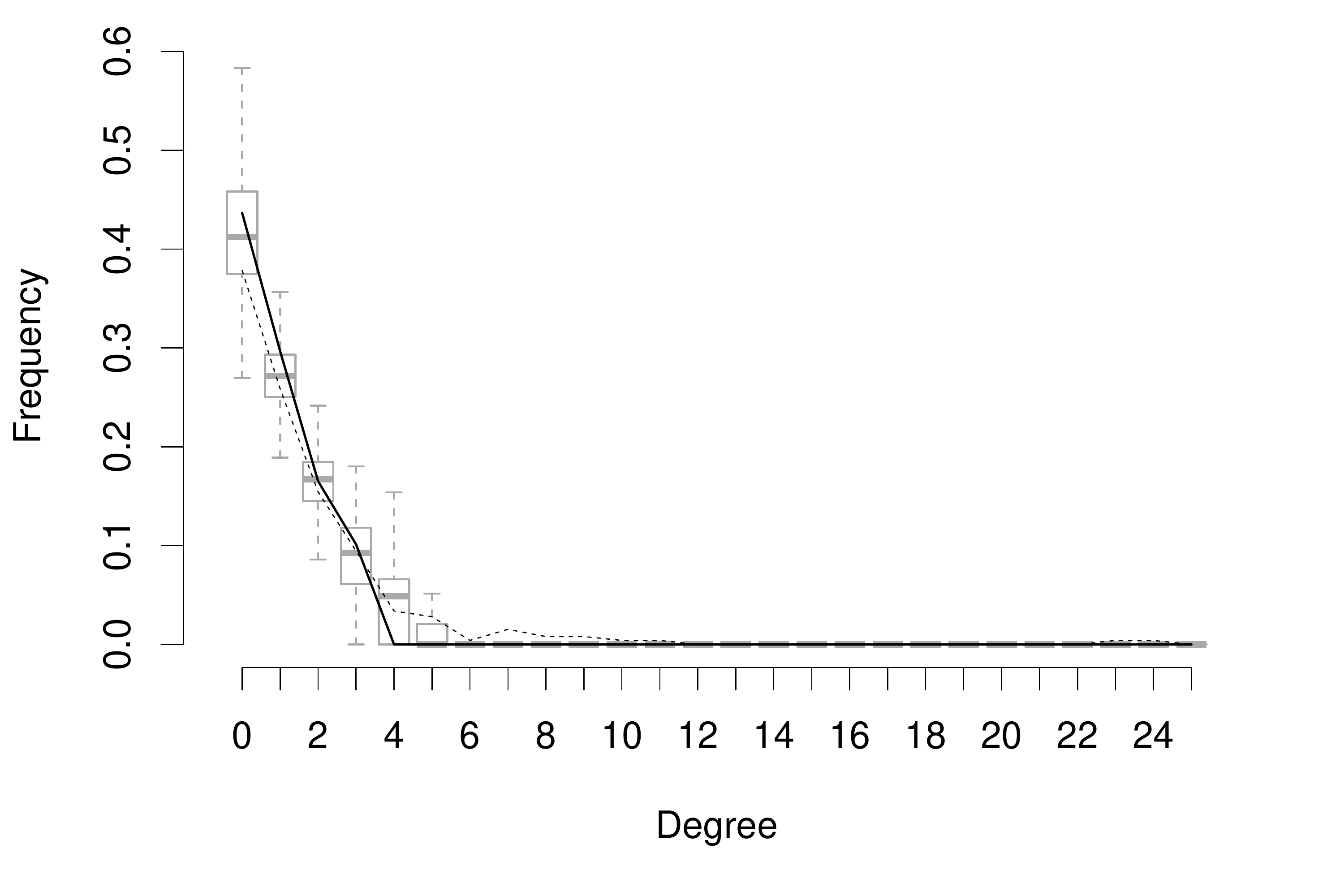} 
\includegraphics[width=.45\textwidth]{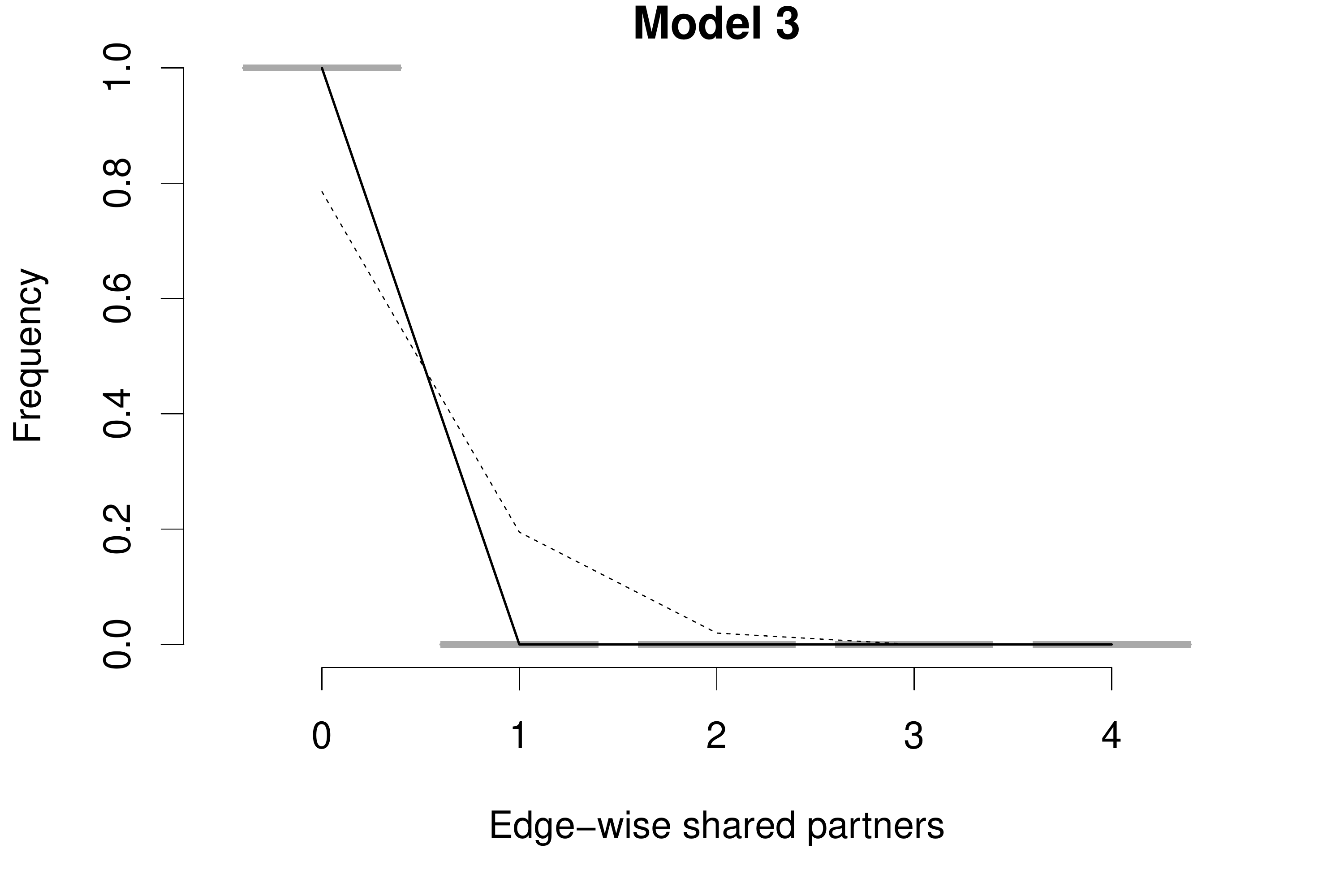} 
\includegraphics[width=.46\textwidth]{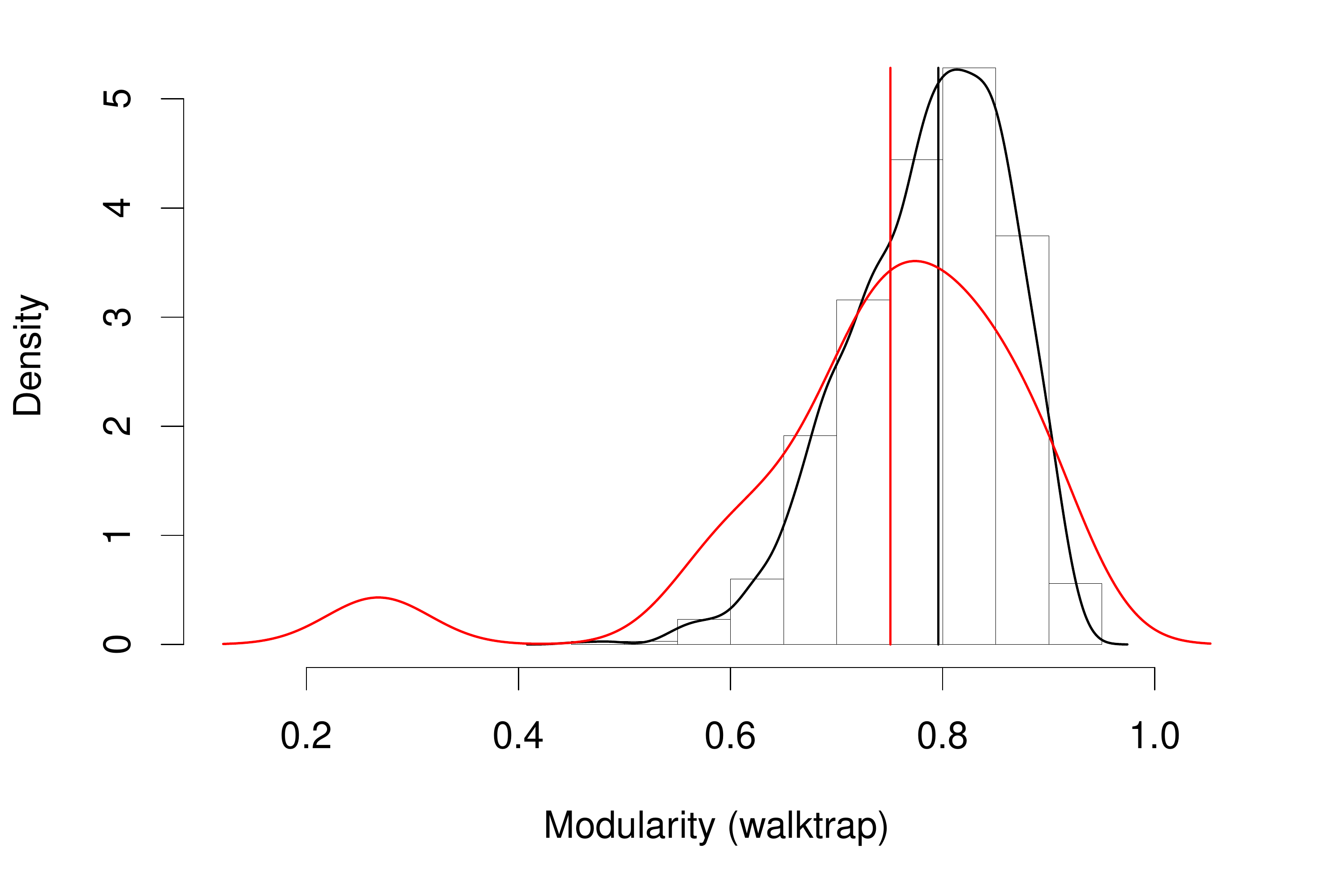} 
\includegraphics[width=.46\textwidth]{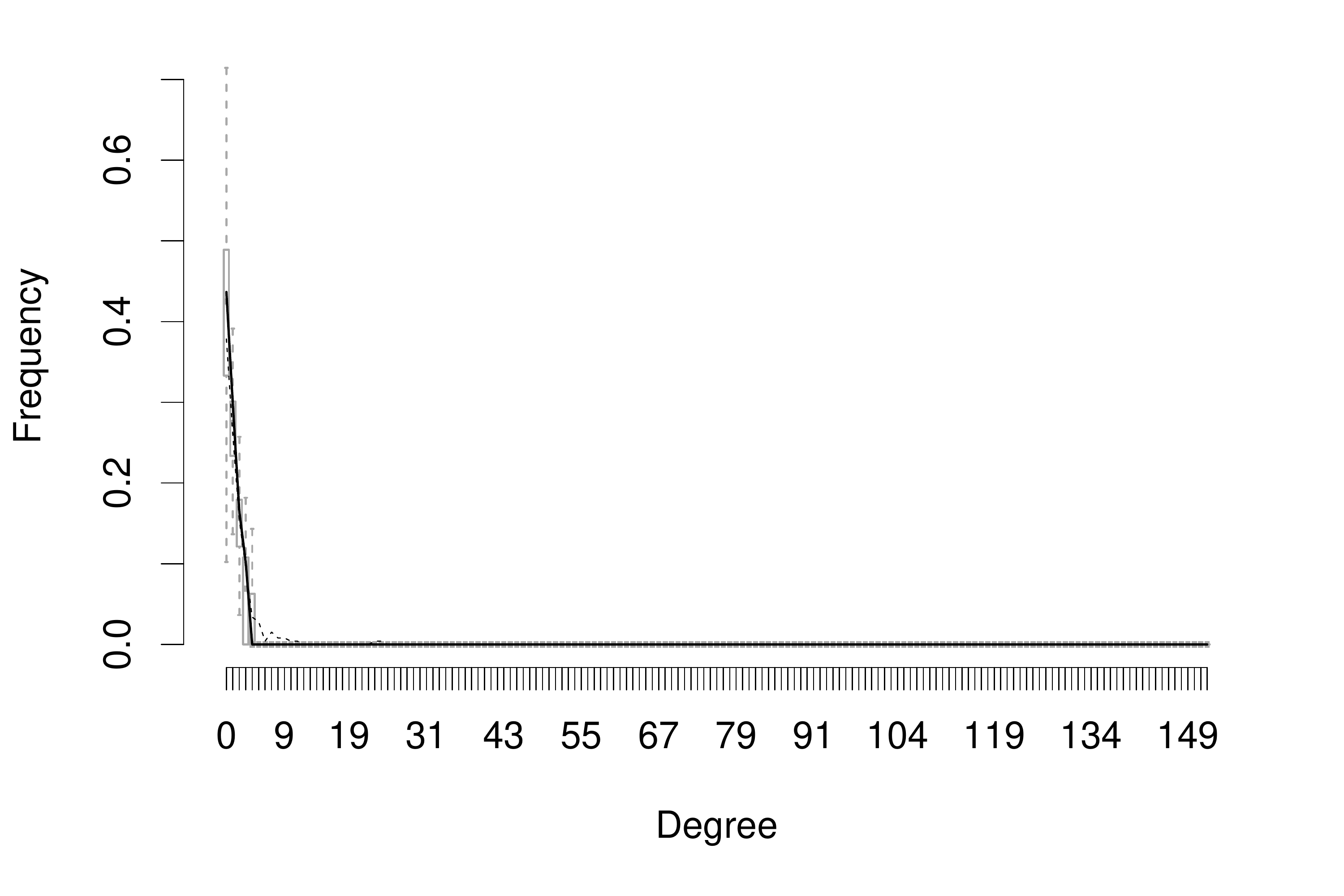} 
\includegraphics[width=.45\textwidth]{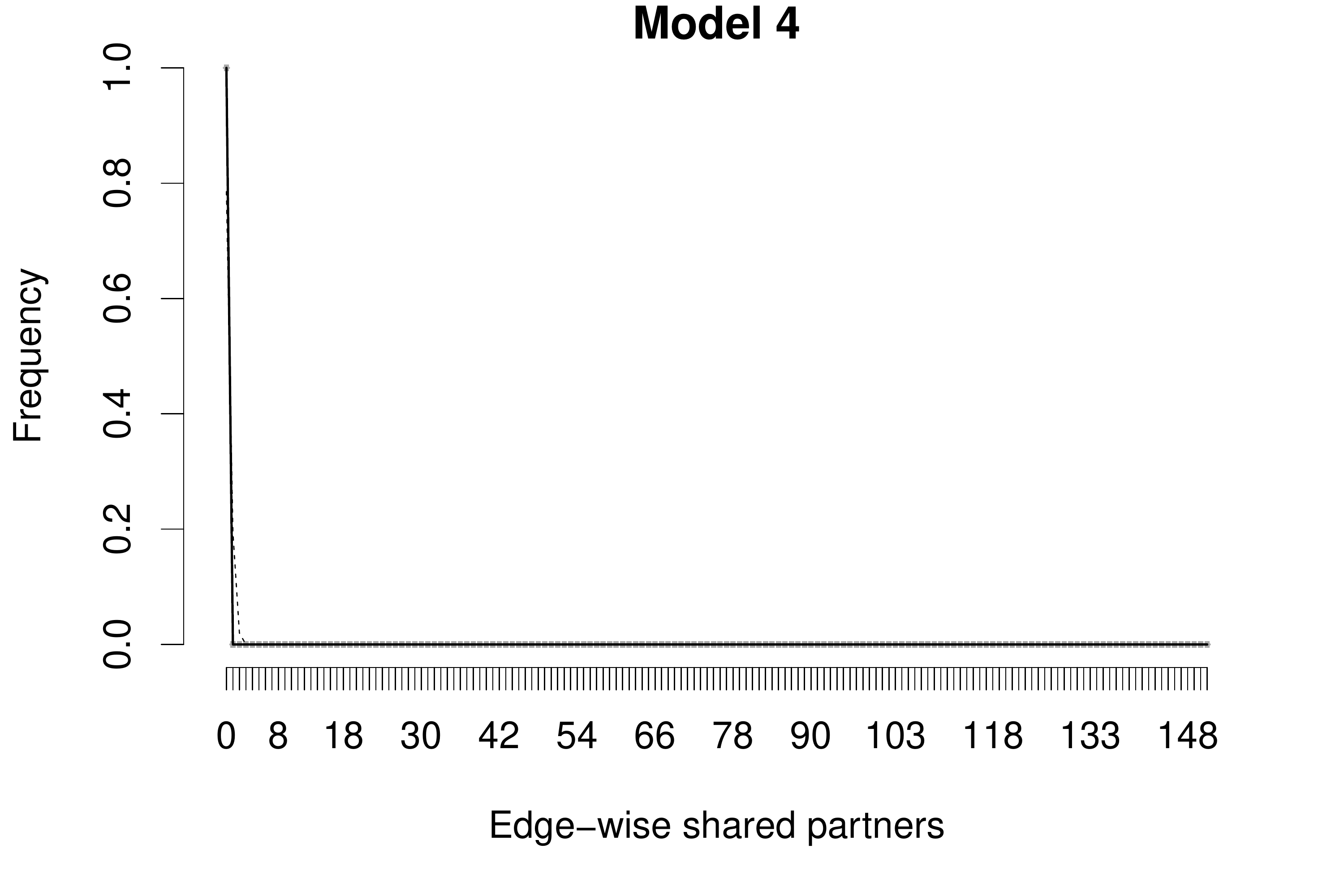} 
\includegraphics[width=.46\textwidth]{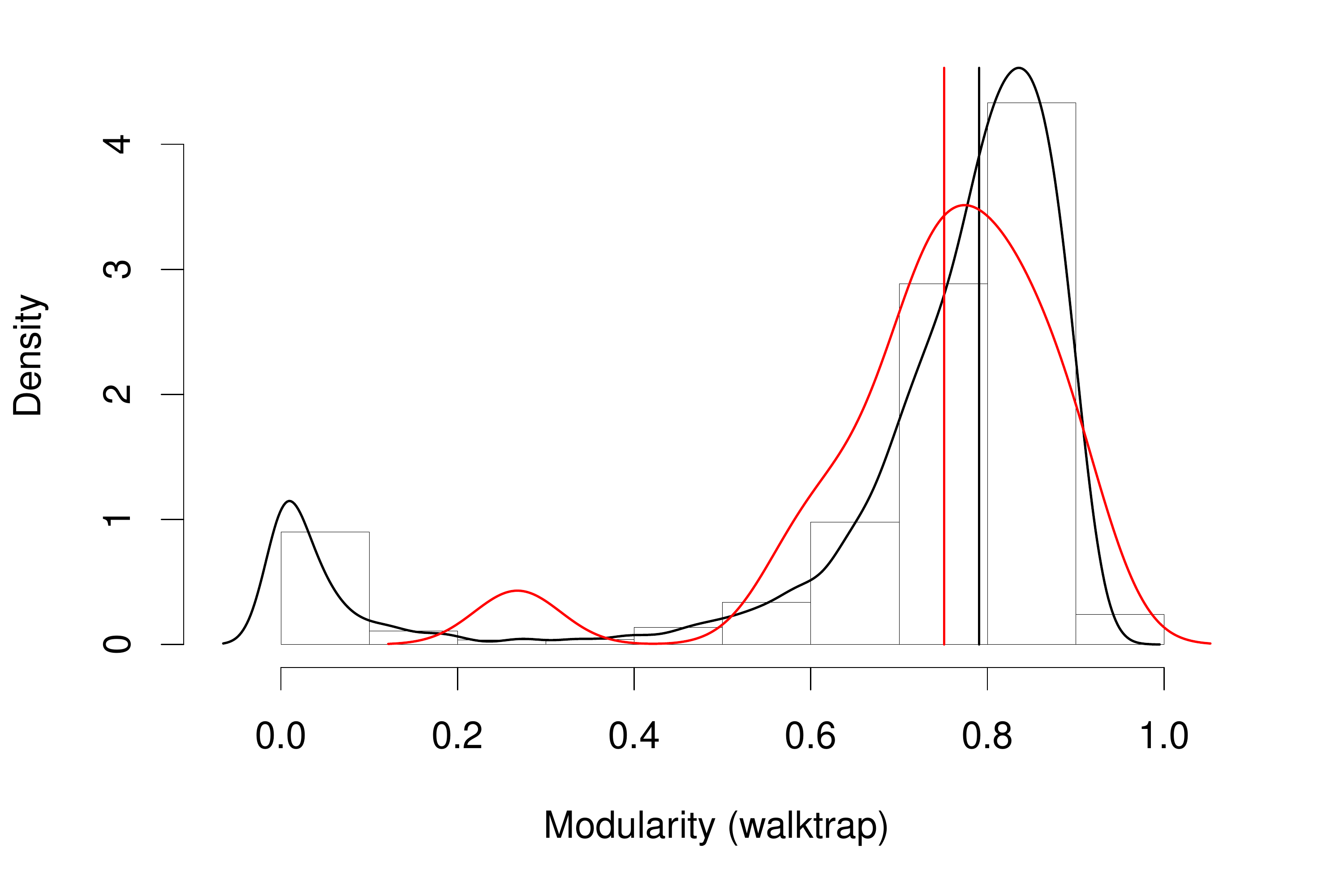} 
\caption{\emph{In-Sample Goodness-of-Fit.} In-sample goodness-of-fit diagnostics as measured by degree, edge-wise shared partners, and modularity, for Model 3 (top row) and Model 4 (bottom row) from the main paper.}
\label{fig:gof2}
\end{figure}
\end{landscape}

\begin{landscape}
\begin{figure}[h!]
\centering
\includegraphics[width=.46\textwidth]{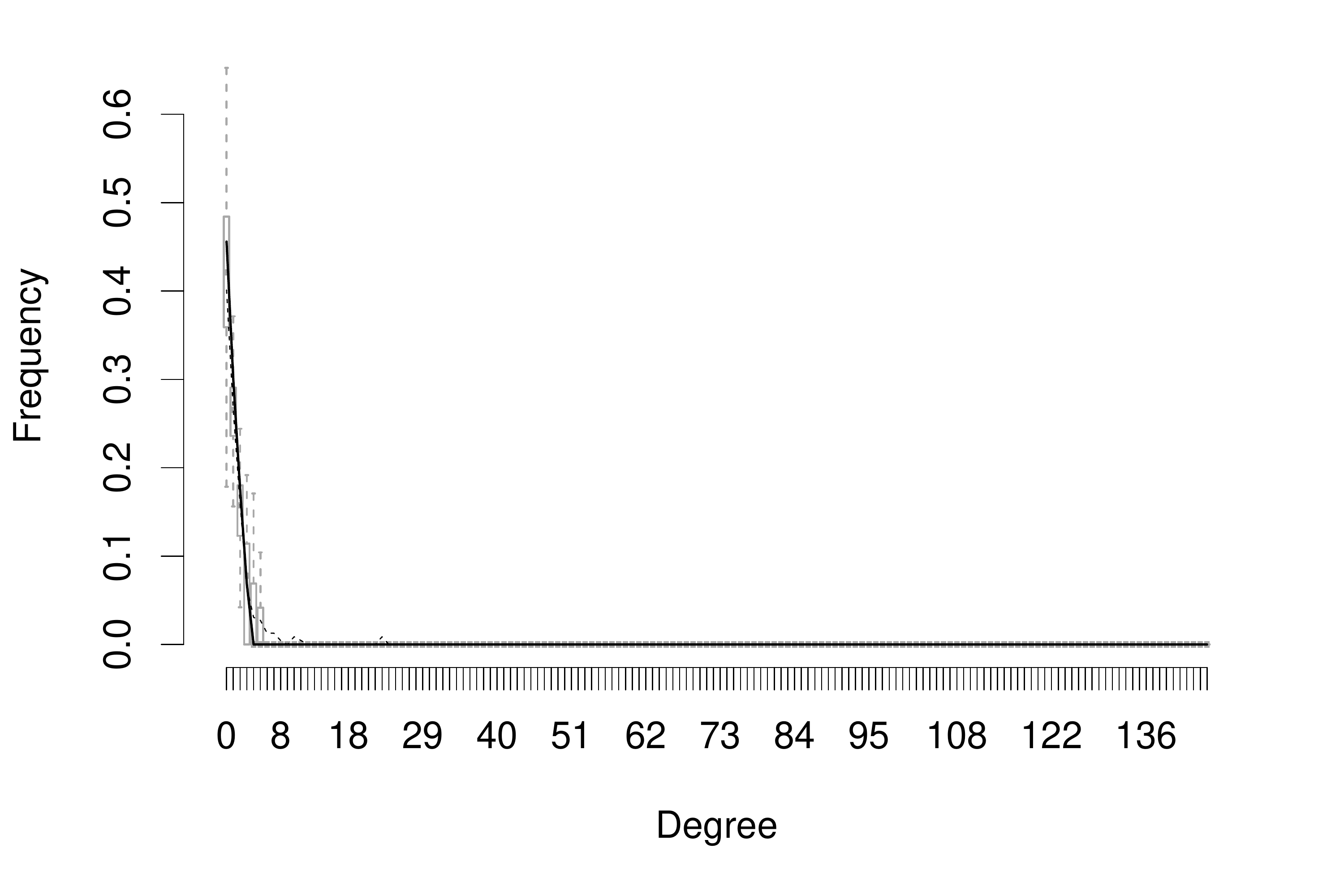} 
\includegraphics[width=.45\textwidth]{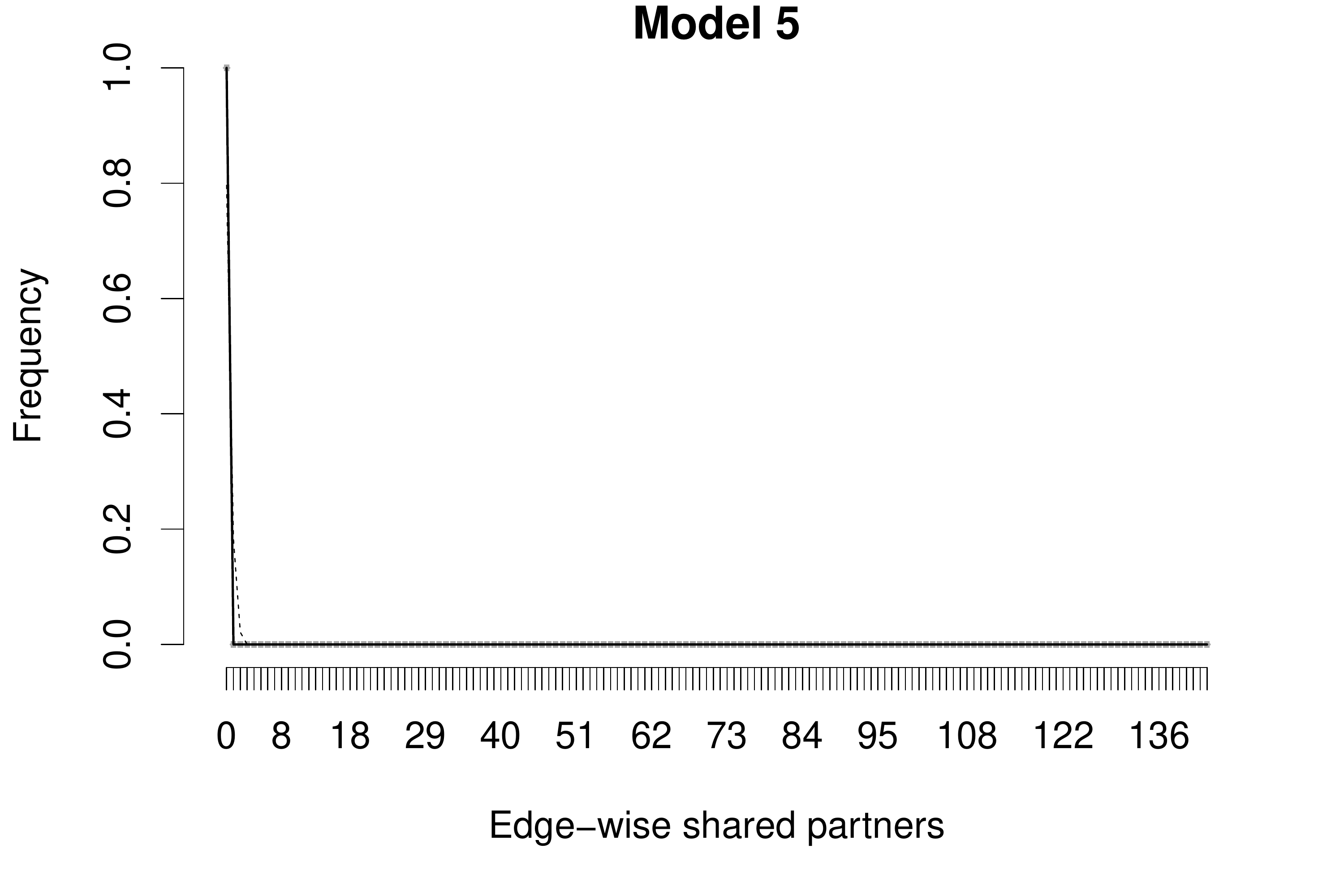} 
\includegraphics[width=.46\textwidth]{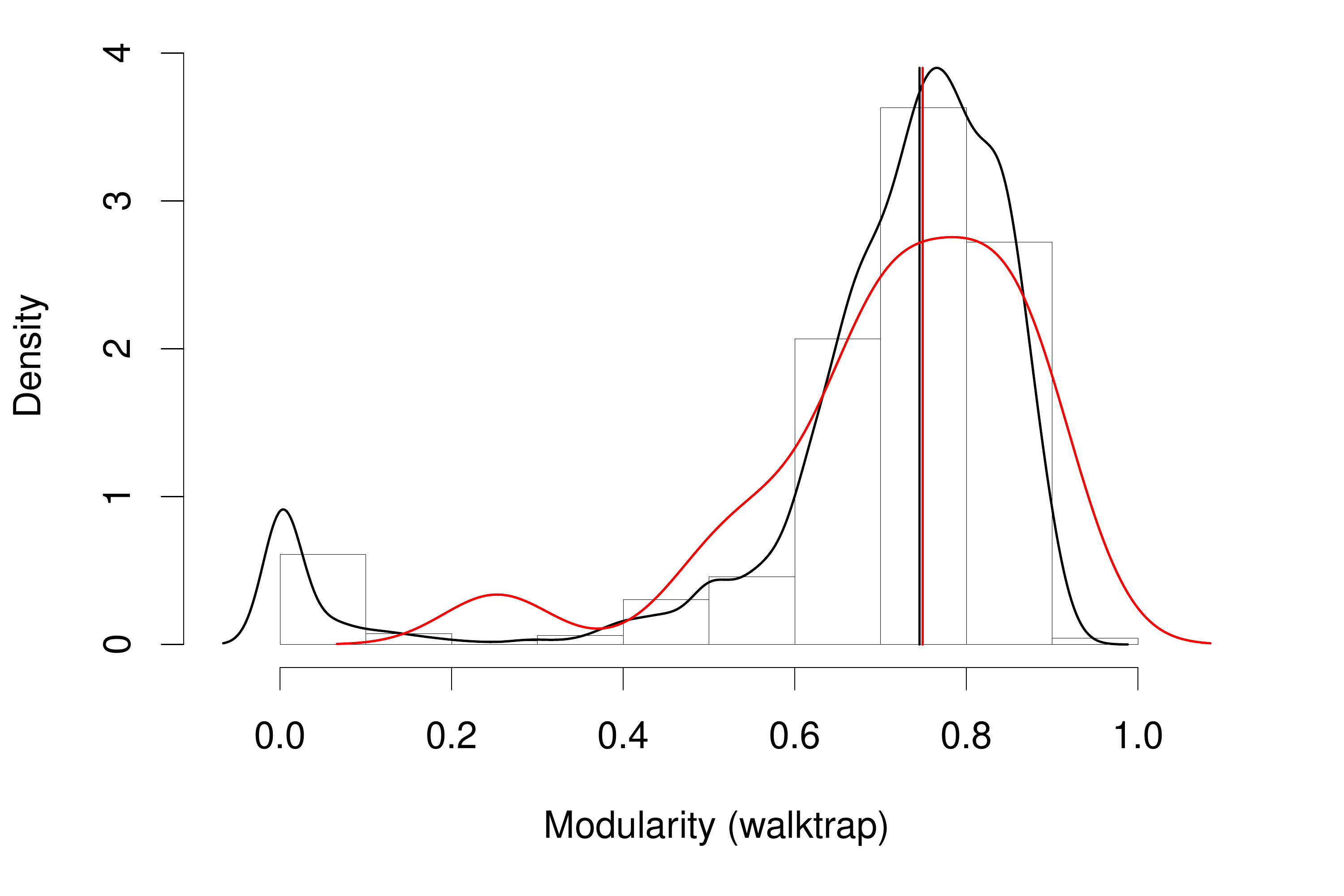} 
\includegraphics[width=.46\textwidth]{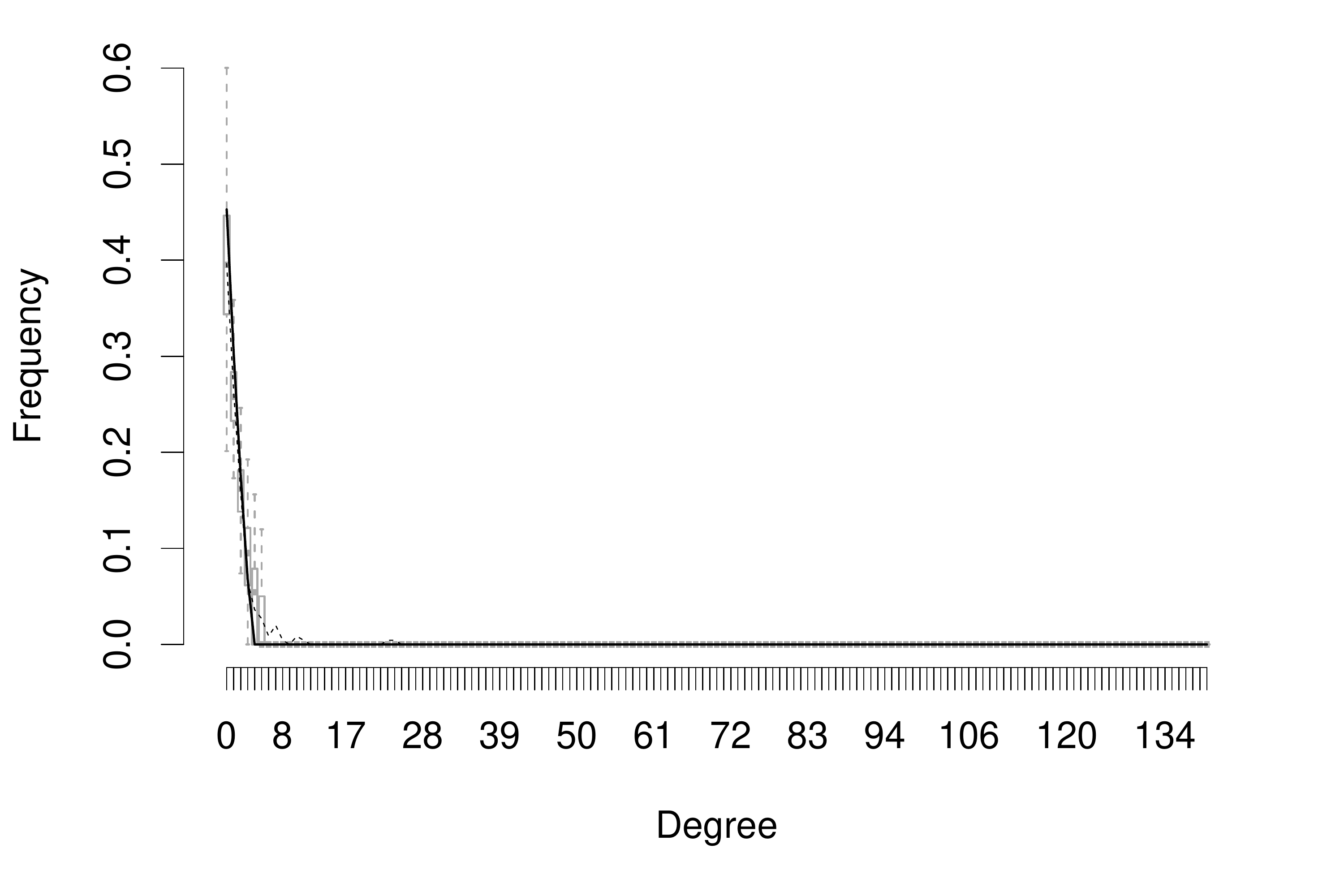} 
\includegraphics[width=.45\textwidth]{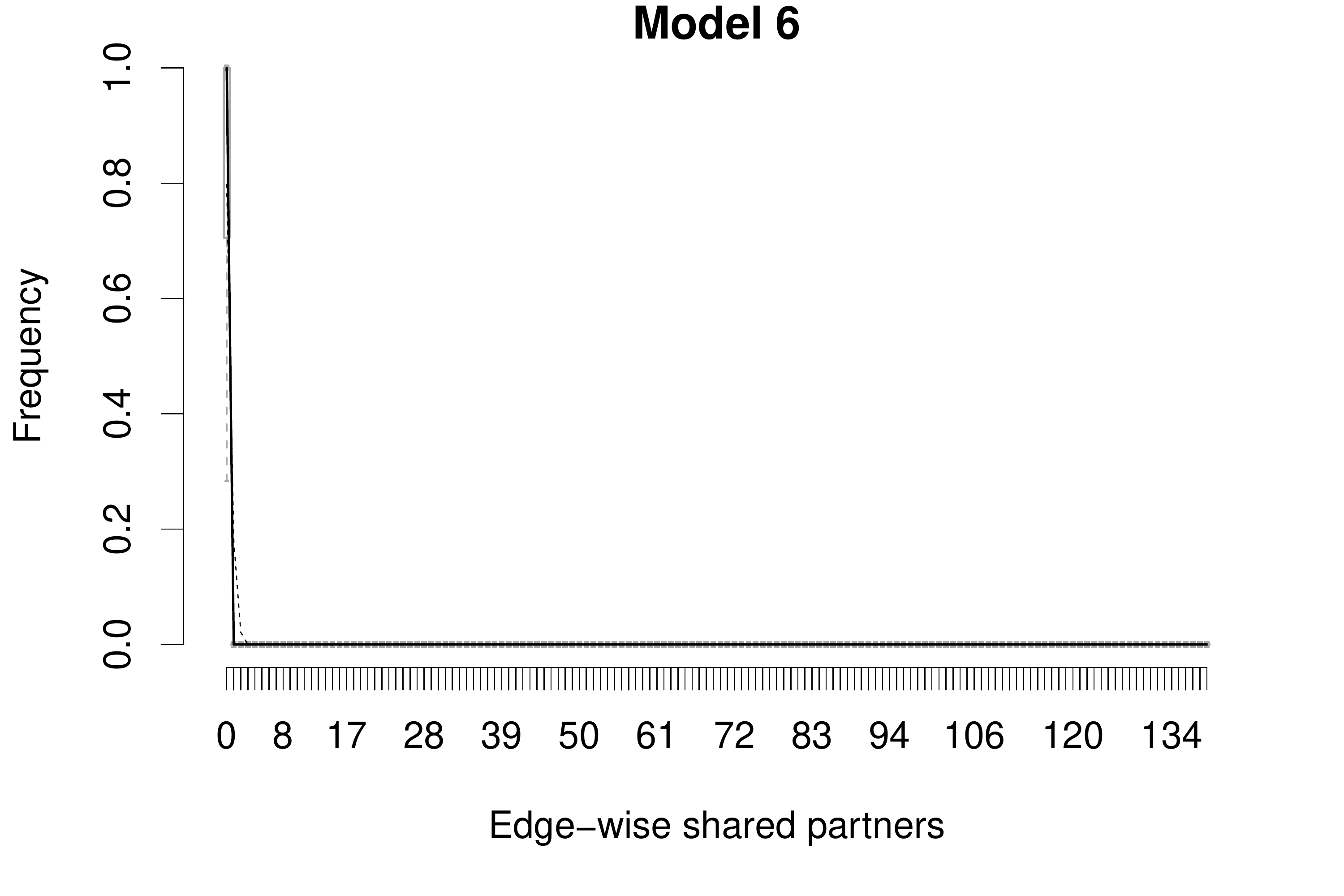} 
\includegraphics[width=.46\textwidth]{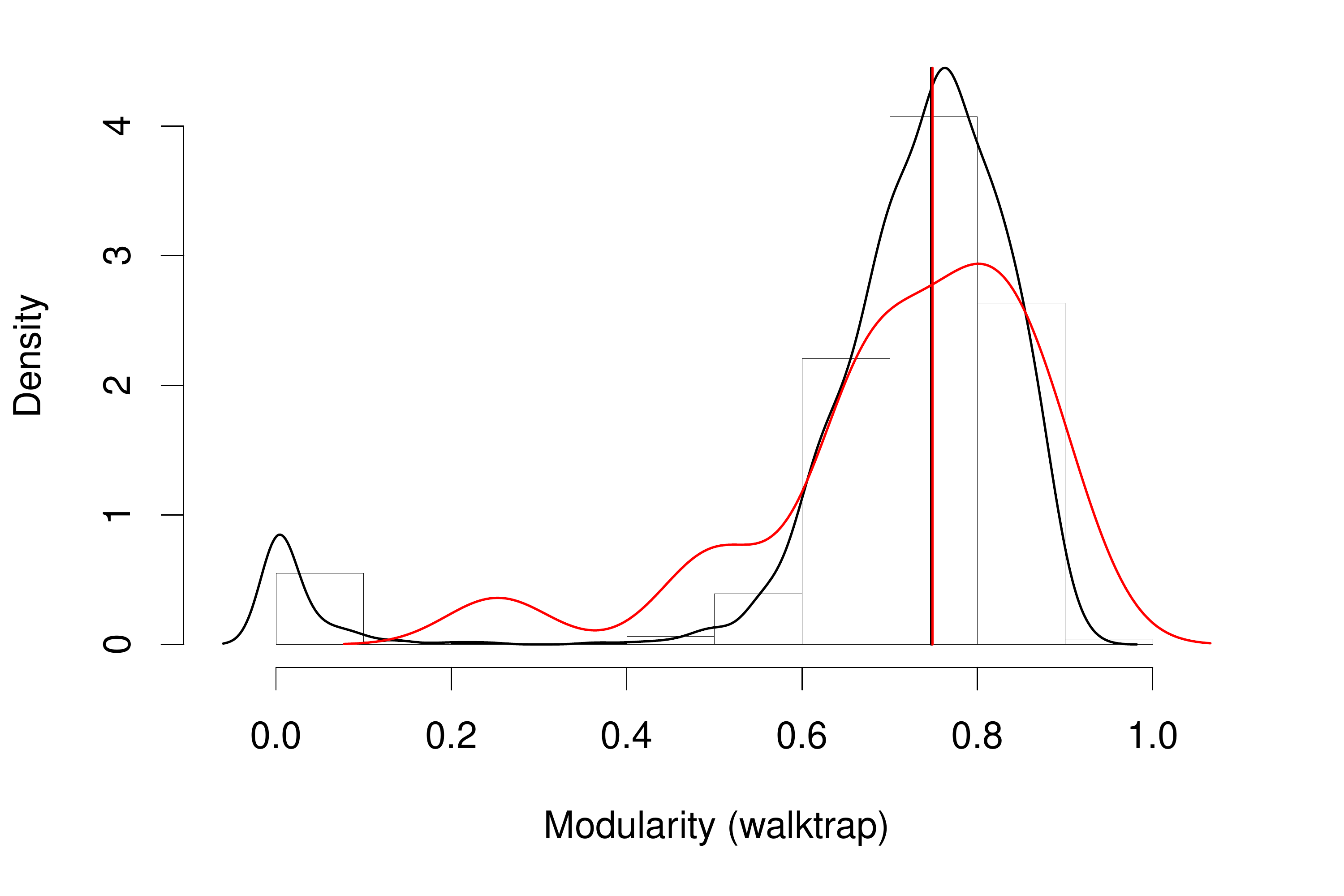} 
\caption{\emph{In-Sample Goodness-of-Fit.} In-sample goodness-of-fit diagnostics as measured by degree, edge-wise shared partners, and modularity, for Model 5 (top row) and Model 6 (bottom row) from the main paper.}
\label{fig:gof3}
\end{figure}
\end{landscape}

In order to compare the fits of the models reported in the main body of the paper, we also present the in-sample fit and out-of-sample performance of each model. For ERGMs, in-sample goodness-of-fit is assessed through the simulation of several networks using the fitted model. Then, the analyst measures how well the simulations capture network statistics that were not originally specified in the model. For each time step, 50 networks were simulated from the fitted model and statistics for degree, edge-wise shared partners, and modularity are plotted below. If the model has sufficiently captured the data generating process (i.e. the onset of violent conflict), then the statistics from the simulated networks (represented as box plots) should be near to the distributions of those statistics in the observed network (represented as black lines). Ideally, the black lines would cross the medians of the simulated box plots. These results are presented in Fig \ref{fig:gof1} through \ref{fig:gof3}, with values for degree and edge-wise shared partners offset by 1 and logged in order to aid in visualization. The simulations from each of the models display quite strong goodness-of-fits. Although neither model drastically outperforms the other, this increases our confidence that the data generating process has been adequately captured. 

\begin{figure}[h!]
\centering
\includegraphics[width=.99\textwidth]{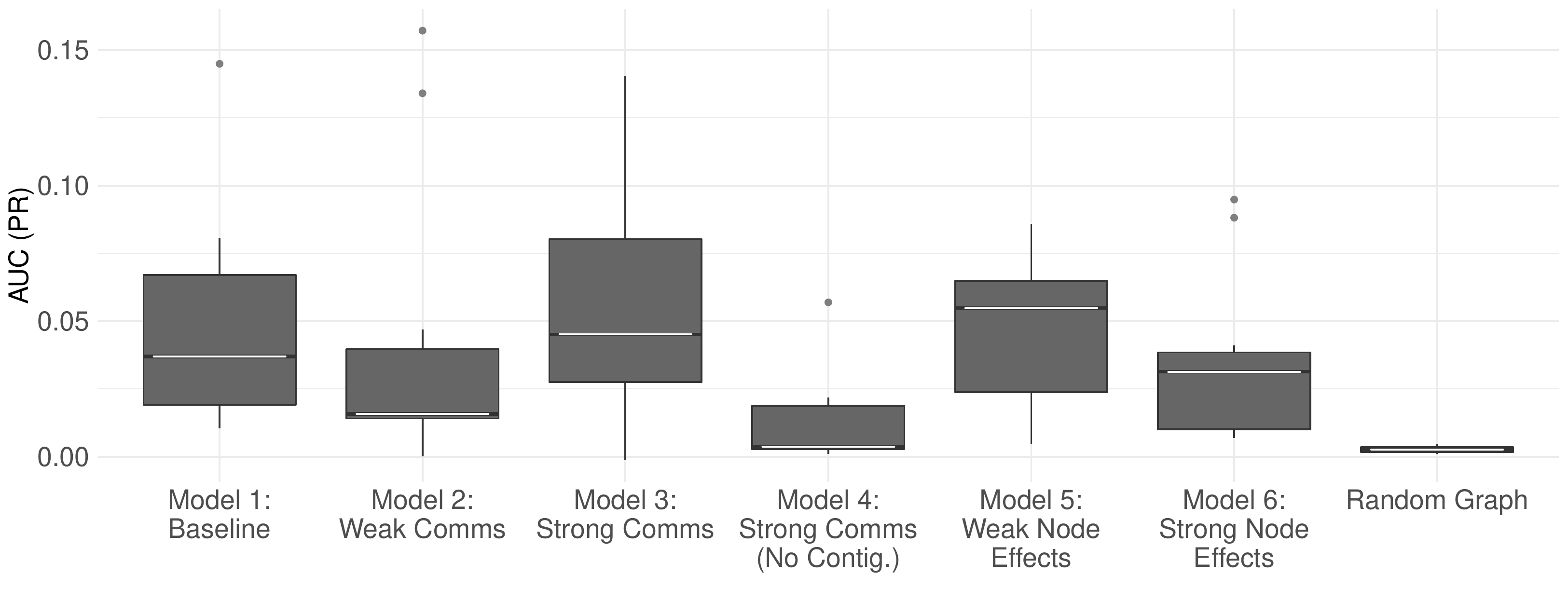} 

\caption{\emph{Test Set Predictive Accuracy.} Out-of-sample predictive performance as area under the precision recall curve for each model in the main paper, as well as the performance of a random graph. Note that Model 5 and Model 6 omit the community bridge effect, because bridges were not present in every iteration, and thus a statistic cannot be calculated for each year.}
\label{fig:box}
\end{figure}

Second, although the prediction of conflict out-of-sample is not an objective of this paper, it is worthwhile to compare the models' relative test set accuracies. This more challenging task helps to assess the extent to which one model over another more adequately captures the data generating process of conflict onset, and addresses the issue of overfitting on the training set. For ERGMs, an especially useful metric is the area under the precision recall curve (AUCPR). For this task, a fitted model is trained on five year windows and the sixth year is taken as a test set for out-of-sample conflict onset prediction (i.e. the formation of a tie in the outcome network of interest). The box plots in Fig \ref{fig:box} display the AUCPR for each of the reported models in the main paper, as well as the performance of a random graph for comparison.

These results suggest that the six models display quite similar out-of-sample predictive performance. Therefore, we find scant evidence for an argument to prefer one model over the other. Instead, the specification likely comes down to theoretical motivations and the research question at hand. Although these model are built for inference, and not prediction, this is further confirmation that conflict onset prediction indeed remains a challenging task for statisticians and political scientists. The in-sample goodness-of-fits, however, indicate that the model is adequately specified to the data at hand.

\clearpage

\begin{table}[ht]
\centering
\begin{tabular}{rllllll}
  \hline
 &\bf war &\bf law &\bf trade &\bf west &\bf human &\bf nuclear \\ 
  \hline
1 & cold & rule & market & east & right & weapon \\ 
  2 & end & norm & commod & north & life & prolifer \\ 
  3 & conflict & principl & product & south & protect & test \\ 
  4 & confront & legal & export & asia & digniti & chemic \\ 
  5 & violenc & intern & industri & bank & valu & destruct \\ 
  6 & fratricid & right & monetari & asian & fundament & arsenal \\ 
  7 & horror & respect & restrict & gaza & justic & arm \\ 
  8 & devast & justic & develop & africa & law & treati \\ 
  9 & destruct & regul & econom & europ & and & disarma \\ 
  10 & after & fundament & system & southern & individu & armament \\ 
   \hline
\end{tabular}
\caption{\emph{Nearest features based on cosine similarity.} Top 10 nearest features in vector space to target feature in column header based on cosine similarity.}
\label{table:nearest}
\end{table}

\clearpage

\clearpage \singlespacing
\bibliographystyle{ieeetr}
\bibliography{multi_comms}

\end{document}